\pdfoutput=1
\documentclass[11pt]{article}

\usepackage[]{acl}

\usepackage{times}
\usepackage{latexsym}
\usepackage{amsmath,graphicx}
\usepackage{soul} % Added
\usepackage{arydshln} % Added
\usepackage{amsfonts} % Added
\usepackage{etoolbox} % Added
\usepackage{booktabs} % Added
\usepackage[noend]{algpseudocode} % Added
\usepackage{algorithmicx,algorithm} % Added
\usepackage{multirow} % Added
\usepackage{amsmath} % Added
\usepackage{bbding} % Added
\usepackage{caption} % Added
\usepackage{subfigure} % Added
\usepackage{color} % Added
\usepackage{url} % Added
\usepackage{hyperref} % Added
\usepackage{graphicx} % Added
\usepackage[bottom]{footmisc} % Added

\newcommand{\xyq}[1]{{\color{blue}{[xyq]}}}
\newcommand{\bee}[1]{{\color{pink}{[surika]}}}
\newcommand{\xlx}[1]{{\color{red}{[xlx]}}}
\newcommand{\pw}[1]{{\color{yellow}{[pw]}}}

\colorlet{soulyellow}{yellow!50}
\colorlet{soullime}{orange!30}
\colorlet{soulblue}{blue!20}
\usepackage{xspace}
\newcommand{\method}{{\textbf{\textsc{Comma}}}\xspace}
\newcommand{\dataset}{{\textbf{\textsc{Hail}}}\xspace}

\usepackage[T1]{fontenc}
\usepackage[utf8]{inputenc}

\usepackage{microtype}

\title{\method: Modeling Relationship among Motivations, Emotions and Actions in Language-based Human Activities}

\author{Yuqiang Xie \quad Yue Hu\footnotemark[2] \quad Wei Peng \quad Guanqun Bi \quad Luxi Xing\\
          Institute of Information Engineering, Chinese Academy of Sciences, Beijing, China \\
          School of Cyber Security, University of Chinese Academy of Sciences, Beijing, China \\
          \texttt{\{xieyuqiang,huyue,pengwei,biguanqun,xingluxi\}@iie.ac.cn} \\}
          
\begin{document}

\maketitle

\begin{abstract}
\footnotetext[2]{Corresponding author.}
Motivations, emotions, and actions are inter-related essential factors in human activities.
While motivations and emotions have long been considered at the core of exploring how people take actions in human activities, there has been relatively little research supporting analyzing the relationship between human mental states and actions.
We present the first study that investigates the viability of modeling motivations, emotions, and actions in language-based human activities, named \method (\textbf{Co}gnitive Fra\textbf{m}ework of Hu\textbf{m}an \textbf{A}ctivities).
Guided by \method, we define three natural language processing tasks (emotion understanding, motivation understanding and conditioned action generation), and build a challenging dataset \textbf{\textsc{Hail}}\footnotemark[3]\footnotetext[3]{We will make our dataset and code publicly available at \url{https://github.com/IndexFziQ/COMMA}.} through automatically extracting samples from Story Commonsense.
% On emotion and motivation understanding tasks, the best model achieved 58.9\%/65.4\% accuracy, far below the human performance of 87.9\%/93.8\%.
% State-of-the-art language generators are more difficult on conditioned action generation task because they lack reasoning abilities that are trivial for humans.
Experimental results on NLP applications prove the effectiveness of modeling the relationship. Furthermore, our models inspired by \method can better reveal the essential relationship among motivations, emotions and actions than existing methods.
\end{abstract}

\section{Introduction}
\label{sec-1}

\begin{figure}[t]
    \centering
    \includegraphics[width=7.5cm]{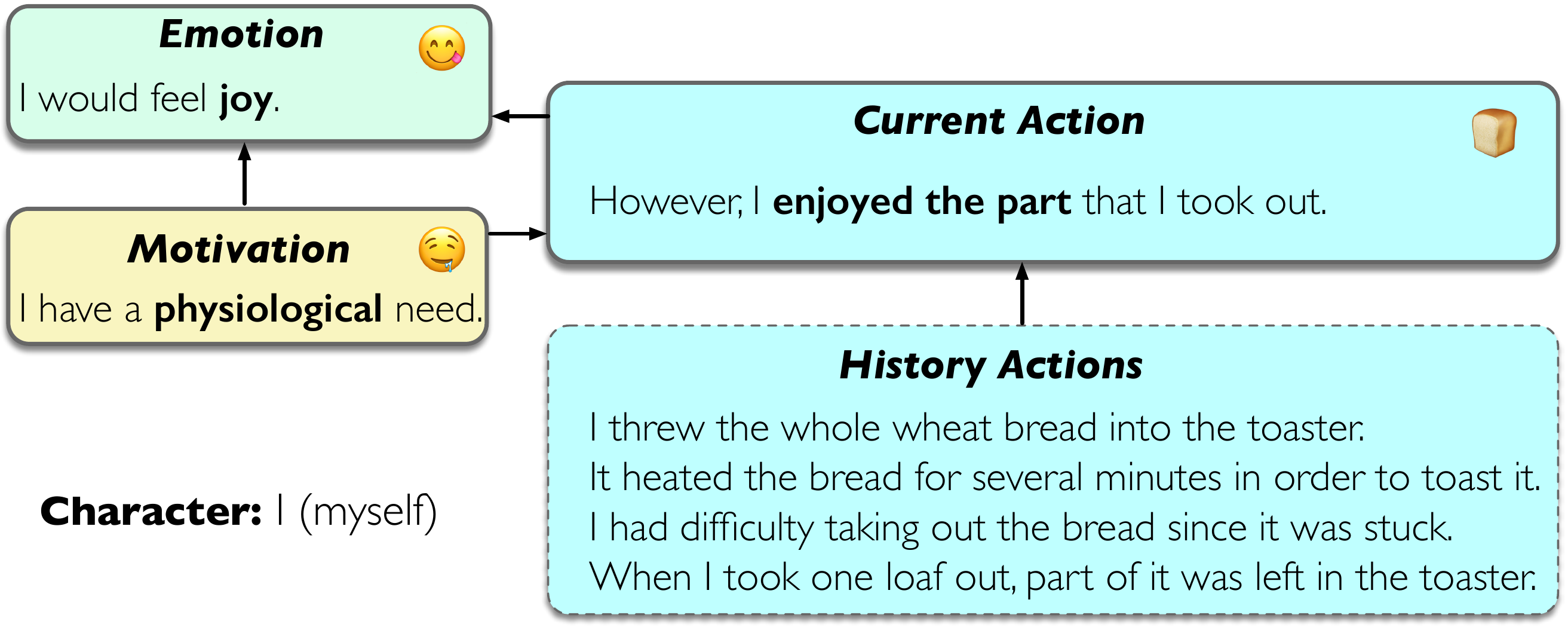}
    \caption{An example of human activity.
    History actions and motivation cause current action. 
    And emotion is the effect of motivation and actions.
    Here, motivation/action/emotion is colored by yellow/blue/green.}
    \label{fig:introduction}
\end{figure}

% useful and interesting problem
Human activities are continuous interactions between external environment (physical world and social events, etc.) and internal mind (motivations, emotions, etc.).
For example, Figure \ref{fig:introduction} demonstrates human activities of character `I' about `eating bread' in external environment, as well as the mental states of `I'.
The motivation of character `I' is a \textbf{physiological} need. 
Conditioned on this motivation and history actions, `I' \textbf{enjoyed the part of bread} on the direction of \textbf{joy} emotion.
While human mental states have long been considered at the core of exploring how people take actions between the lines in language-based human activities, there has been relatively little research supporting analyzing the relationship between human mental states and actions.
It is challenging to comprehensively model the relationship of motivations, emotions and actions in language-based human activities, which can allow researchers to reason the essential causes of human activities from the cognitive perspective and supply reasonable explanations.
This technology will have a profound impact on various natural language processing (NLP) downstream applications, such as intelligent dialogue, controllable text generation, recommendation systems, and public opinion analysis.

% background and unsolved problem
In recent years, traditional sentiment analysis technology has been widely used \cite{DBLP:conf/emnlp/SocherPWCMNP13,DBLP:conf/emnlp/HamiltonCLJ16}, which mainly focuses on sentiment detection.
Although the current state-of-the-art sentiment analysis system can detect the polarity of text \cite{DBLP:journals/widm/ZhangWL18} or consider fine-grained categories (a.k.a. aspects) to make predictions \cite{DBLP:conf/semeval/PontikiGPAMAAZQ16}, the analysis of predictions and interpretations of its causes are still limited.
Lately, a large amount of work introduces human motivations into sentiment analysis and action analysis \cite{DBLP:conf/acl/KnightCSRB18,DBLP:conf/acl/SmithCSRA18,DBLP:conf/aaai/SapBABLRRSC19,DBLP:journals/corr/abs-1904-09728,Wei2022CogIntAc}.
However, the aforementioned works focus on the analysis of the relationship between ``motivations and actions'' or ``emotions and actions'', without modeling a unified consideration of the relationship among motivations, emotions and actions.

% related theory
Researches on human activities have increased over the past two decades with many fields contributing including psychology, computer science and so on.
In this paper, we focus on works in two human mental states, motivations and emotions, that drive human activities.
As for motivation, psychologist Hull \cite{Hull1974EssentialsOB} believes that motivation is the drive for human actions and explains why people initiate, continue or terminate a certain action at a particular time.
% After that, many researches discuss the topic ``how the forces which determine action arise''.
% Recently, \cite{Kruglanski2014FromRT} create a new concept of \textit{Motivational Readiness} to address this problem which generally model how motivation works from readiness to action, where the key elements of motivation is want and expectancy.
From area of emotion, numerous theories \cite{emotion-annurev,Kagan2007WhatIE,Smith2016TheBO} that attempt to explain the origin, function, and other aspects of emotions have fostered more intense research on emotion topic.
% However, there has not been reached scientific consensus on the definition of emotion.
Psychologist Plutchik \cite{Plutchik1980AGP} establishes a general psycho-evolutionary theory of emotion, which introduces eight specific distinct basic emotions. 
Each of basic emotions represents adaptation to a prototypical task in human activity.

Aiming at modeling the relationship among human motivations, emotions and actions in language-based human individual activities, we propose a general \textbf{Co}gnitive Fra\textbf{m}ework of Hu\textbf{m}an \textbf{A}ctivities (\method).
% More specifically, we define basic elements of human activities and model the relationships by analyzing where are actions/emotions from.
These relationships will help the researchers of NLP areas track the cause of people's emotions and actions, and give a more reasonable explanation and analysis for results.
To verify the effectiveness of our framework, we propose three NLP understanding and generation tasks, including emotion understanding, motivation understanding, and conditioned action generation.
More concretely, we construct a new dataset \textbf{\textsc{Hail}} (\textbf{H}uman \textbf{A}ctivities \textbf{I}n \textbf{L}ife) by automatically extracting samples with complete mental state annotation from Story Commonsense \cite{DBLP:conf/acl/KnightCSRB18}.
Experimental results on NLP applications prove the effectiveness of modeling the relationship. Furthermore, our models inspired by \method can better reveal the essential relationship among motivations, emotions and actions than existing methods.

\section{Background}
\label{sec-3}

Human activities are interactions between internal mind of people and the external environment.
In this part, we will describe the basic elements of \method and modeling the relationship among elements in details.

\begin{figure}[ht]
    \centering
    \includegraphics[width=0.28\textwidth]{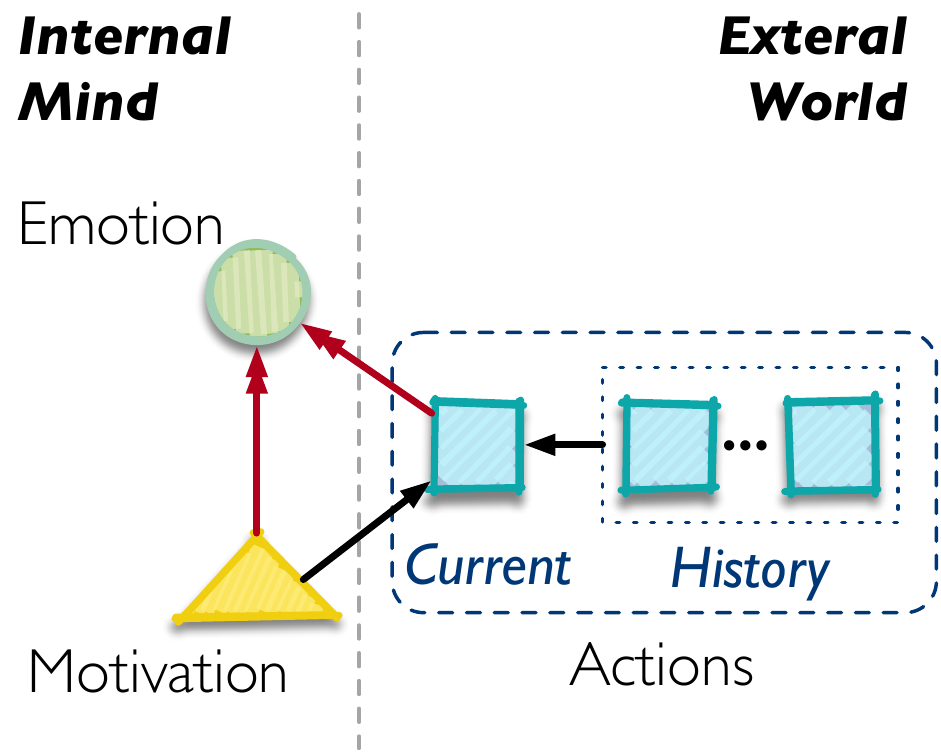}
    \caption{Relationship Modeling in \method.}
    \label{fig:framework}
\end{figure}

% The basic elements of human activities in \method include motivations, emotions and actions.

\begin{figure*}[ht]
\centering                                 
    \subfigure[\textit{Emotion Understanding}]{         
    \begin{minipage}{5cm}
        \centering            
        \includegraphics[width=4cm]{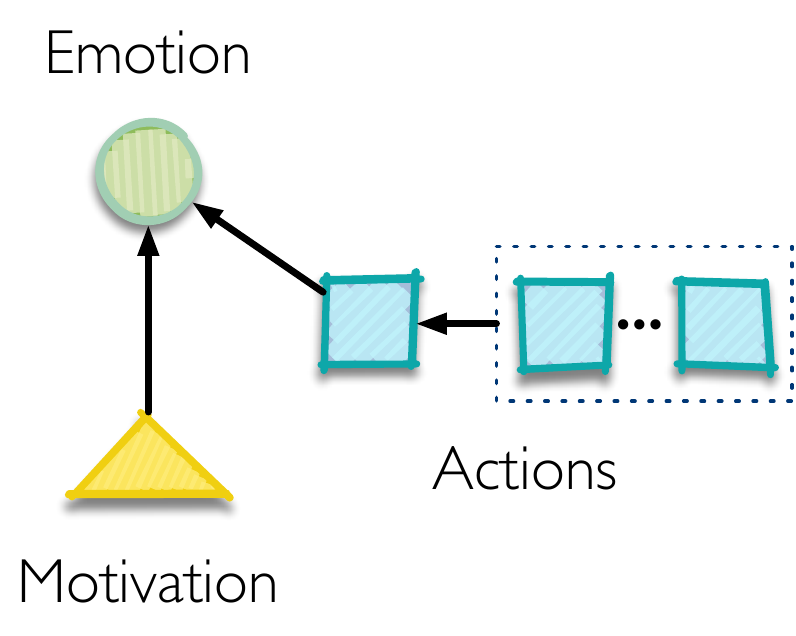}  
        \label{fig:relationship-ep}
    \end{minipage}}
    \subfigure[\textit{Motivation Understanding}]{         
    \begin{minipage}{5cm}
        \centering
        \includegraphics[width=4cm]{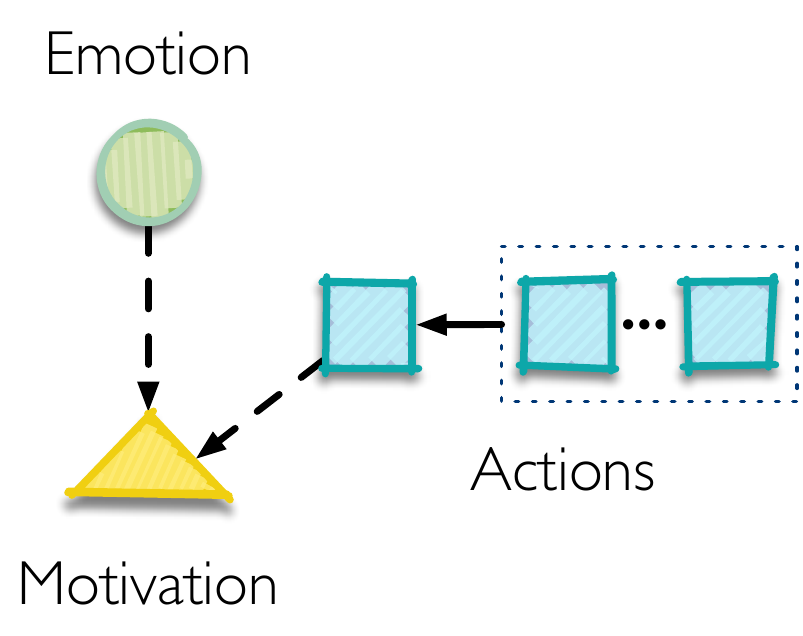}
        \label{fig:relationship-mp}
    \end{minipage}}
    \subfigure[\textit{Conditioned Action Generation}]{         
        \begin{minipage}{5cm}
        \centering            
        \includegraphics[width=4cm]{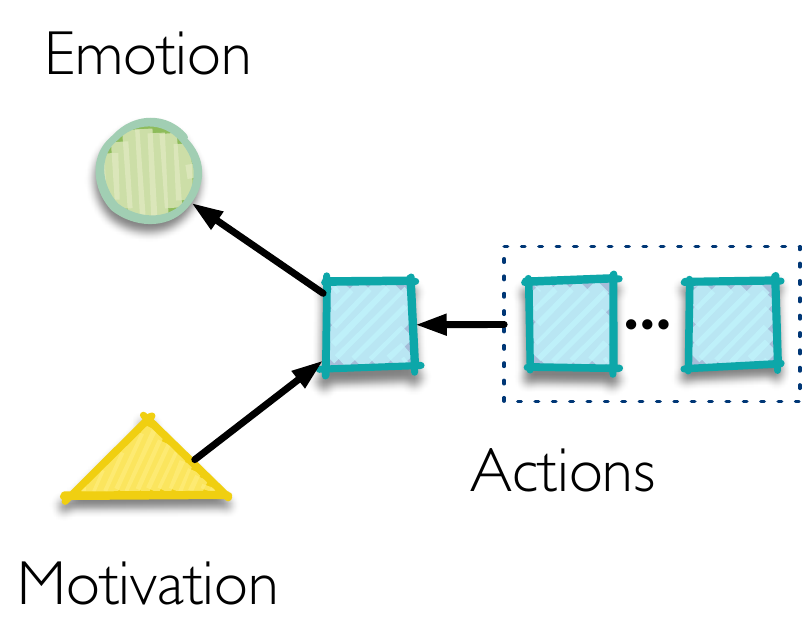}  
    \label{fig:relationship-ap}
    \end{minipage}}
\caption{Three tasks for modeling motivations, emotions, and actions in language-based human activities.
The solid line and the dotted line represent forward and reverse reasoning respectively.}
\label{fig:relationship}
\end{figure*}

\subsection{Basic Elements in \method}
\label{sec-3-2}
\noindent \textbf{Motivations} are the innate physical or psychological motivations of human beings which are the origin of human activities. Different psychological theories have different classification rules for human motivations. We utilize \textit{hierarchy of needs} of \cite{Maslow2013ATO} (\textit{\textbf{physiological needs, stability, love and belonging, esteem, self-actualization}}).
% as illustrated in Fig. \ref{fig:mbe}.

\noindent \textbf{Emotions} are the psychological responses of motivations to the degree of satisfaction with the external environment. 
% It is formed by motivations (internal causes) and external environments (external causes). 
We employ the \textit{wheel of emotions} of \cite{Plutchik1980AGP} and use eight basic emotional dimensions (\textit{\textbf{joy, trust, sadness, surprise, fear, disgust, anger, and anticipation}}).
% as illustrated in Fig. \ref{fig:mbe}. 
It has become a common choice in the existing emotion categorization literature \cite{DBLP:journals/ci/MohammadT13,DBLP:conf/emnlp/ZhouZZZG16,DBLP:conf/acl/KnightCSRB18}.

\noindent \textbf{Actions} are people's actions that interact with the external environment. 
Actions are under the psychological condition of ``one has a certain need and develops on the direction of future emotion'' in this paper.
Limited to the annotations of our based data Story Commonsense \cite{DBLP:conf/acl/KnightCSRB18}, we treat story events as actions, which is a sentence in language-based form. 
That is, \textit{\textbf{a story event equals to an action}} in this paper.

\subsection{Relationship Modeling}
\label{sec-3-4}

As demonstrate in Fig. \ref{fig:framework}, \method is composed of internal mind and external environment, where people own internal mind (motivations, emotions) and take actions in external environment. 
The solid line and the dotted line represent forward and reverse reasoning respectively.
We will attempt to model relationships among the basic elements by answering the next two questions. 

\noindent \textbf{{Q1: Where are Actions from? }}
Following the view of Hull \cite{Hull1974EssentialsOB}, motivation is the drive for human actions. 
Intuitively, current action also caused by history actions. 
As shown in Fig. \ref{fig:framework}, motivation and history actions leads to the development of current action together. 
Meanwhile, actions develop in the direction of the future emotion.
For instance, one wanted to eat, and one could eat some food and felt happy then. 
Conversely, one could have nothing to eat and felt sad.

\noindent \textbf{{Q2: Where are Emotions from? }}
Emotions are mental states brought on by neurophysiological changes, variously associated with thoughts, feelings, behavioural responses \cite{Panksepp1998AffectiveNT,Cabanac2002WhatIE}. Simplicity, emotions comes from actions in external world and other complex mental states. In this work, we predigest this complicated process. 
As demonstrated in Fig. \ref{fig:framework},
emotion is conditioned by whether the action satisfy the primary motivation. 
E.g., one wanted to eat, and one would happy if he ate some food, either sad if no restaurant opened.

All in all, relationships of motivations, emotions and actions are demonstrated in Fig. \ref{fig:framework}:

\noindent \textbf{(1) Motivation and history actions cause action;}

\noindent \textbf{(2) Emotion is \textbf{effect} of motivation and action.}

These relationships will help the researchers track human's motivations, emotions and actions. 
% In addition, a more reasonable analysis for human activities will be presented.

\section{Tasks and Data}
\label{sec-4}

To verify the effectiveness of \method, we propose three language-based understanding and generation tasks, including emotion understanding, motivation understanding, and conditioned action generation.
Correspondingly, we build a \textbf{\textsc{Hail}} dataset by automatically extracting samples with complete mental state annotation from Story Commonsense \cite{DBLP:conf/acl/KnightCSRB18}.
% We will describe tasks and data collection in the following.

\begin{figure*}[ht]
    \centering
    \includegraphics[width=0.9\textwidth]{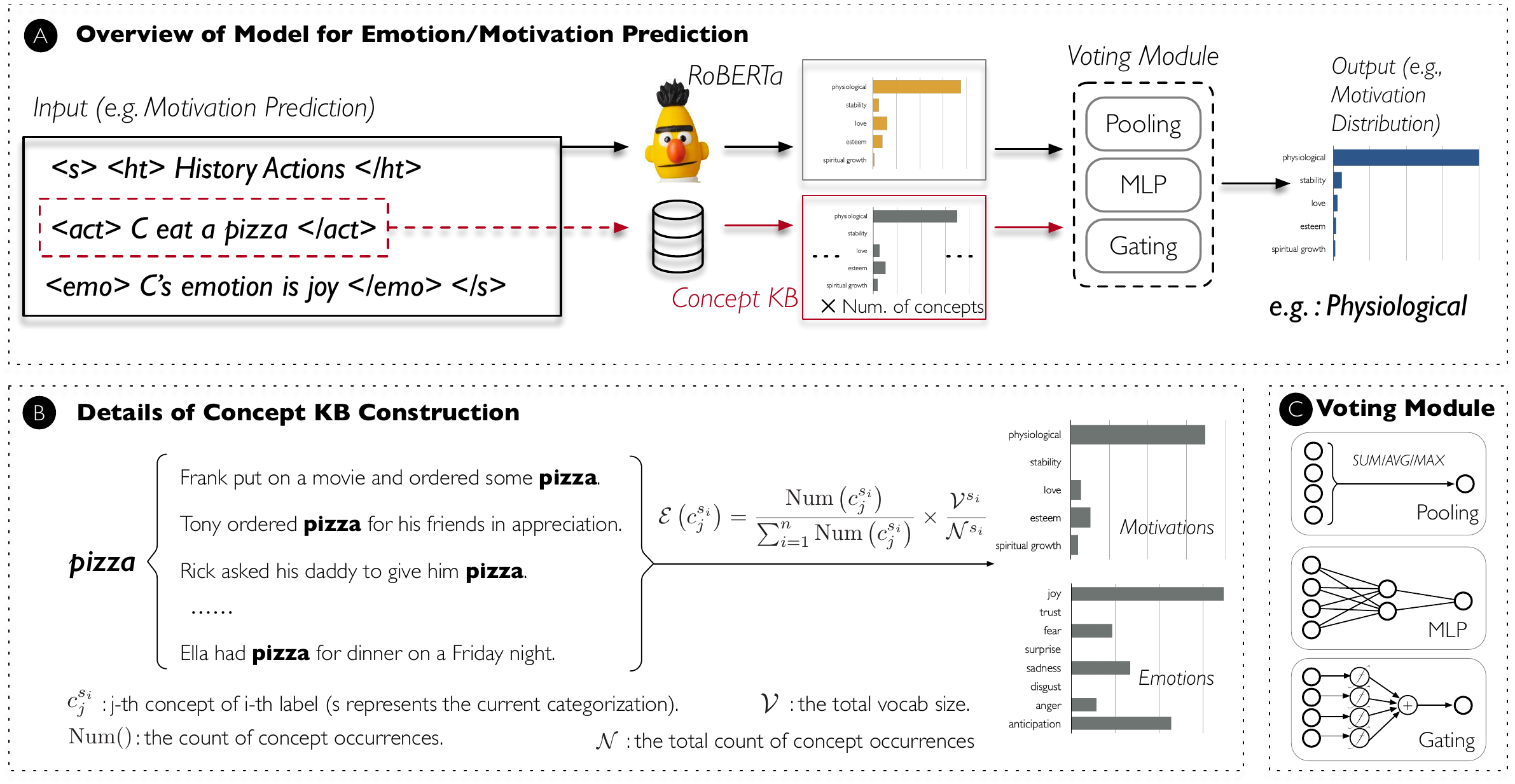}
    \caption{Model overview (A) for emotion and motivation understanding tasks (emotion understanding as an example). B is the detail of concept knowledge base construction. C shows three options of voting module in A.}
    \label{fig:model_cls}
\end{figure*}

\subsection{Task Definition}

\noindent\textbf{Emotion Understanding (EU)}
We formulate emotion understanding as sequence classification problems consisting of history actions, current action and people's motivations as context and a objective resulted emotion. Each instance in \dataset is defined as follows:

\noindent $\mathcal{A}$: The current action.

\noindent $\mathcal{H}$: The history actions.

\noindent $\mathcal{C}$: The character of current action.

\noindent $\mathcal{M}$: The motivation to drive the current action.

\noindent $\mathcal{E}$: The resulted emotion of the current action.

As shown in Fig. \ref{fig:relationship-ep}, given the motivation $\mathcal{M}$, character $\mathcal{C}$ and all actions ($\mathcal{H}$ and $\mathcal{A}$), the EU task is to select the most plausible emotion $\mathcal{E}$.
% $P(\mathcal{E}|\mathcal{H}, \mathcal{A}, \mathcal{M})$.

\noindent\textbf{Motivation Understanding (MU)} 
As is shown in Fig. \ref{fig:relationship-mp}, compared with EU, motivation understanding is a reverse reasoning process.
Given the emotion $\mathcal{E}$, character $\mathcal{C}$ and all actions ($\mathcal{H}$ and $\mathcal{A}$), the MU task is to reversely reason about the most plausible motivation $\mathcal{M}$.
% $P(\mathcal{M}|\mathcal{H}, \mathcal{A}, \mathcal{E})$.

\noindent\textbf{Conditioned Action Generation (CAG)} 
As shown in Fig. \ref{fig:relationship-ap}), CAG is the task of generating a valid action $\mathcal{A}$ conditioned on the history actions $\mathcal{H}$, character $\mathcal{C}$, motivation $\mathcal{M}$ and emotion $\mathcal{E}$. Formally, the task requires to maximize $P(\mathcal{A}, \mathcal{E}|\mathcal{H}, \mathcal{C}, \mathcal{M})$.

\begin{table}[ht]
    \centering
    \scalebox{0.75}{
    \begin{tabular}{cccccc}
    \toprule[1pt]
    \multicolumn{2}{c}{\bf EU Task} & \multicolumn{2}{c}{\bf MU Task} & \multicolumn{2}{c}{\bf CAG Task} \\
    \toprule[0.5pt]
    Input & output & Input & output & Input & output \\
    \toprule[0.5pt]
    $\mathcal{H}$,$\mathcal{A}$,$\mathcal{C}$,$\mathcal{M}$   
    &  $\mathcal{E}$  
    & $\mathcal{H}$,$\mathcal{A}$,$\mathcal{C}$,$\mathcal{E}$ 
    & $\mathcal{M}$ 
    & $\mathcal{H}$,$\mathcal{C}$,$\mathcal{M}$ 
    & $\mathcal{A}$,$\mathcal{E}$\\
    \toprule[1pt]
    \end{tabular}
    }
    \caption{Input and Output of tasks in \dataset.}
    \label{tab:task-definition}
\end{table}
% \subsection{Data Collection}

% We automatically collect a new dataset \textbf{\textsc{Hail}} for the above three tasks by extracting from the existing resource, Story Commonsense\cite{DBLP:conf/acl/KnightCSRB18}, with the help of NLP tools. 
% In all, we extract 13,568 examples in Story Commonsense that meet our requirements. The I/O are summarized in Table \ref{tab:task-definition}. 
% More details, refer to Appendix \ref{appdix-data}.

\subsection{Data Collection}
% \label{appdix-data}

To verify the effectiveness of our cognitive framework, we construct a new dataset \textbf{\textsc{Hail}} (\textbf{H}uman \textbf{A}ctivities \textbf{I}n \textbf{L}ife) for the above four tasks by automatically extracting from the existing resource, Story Commonsense\cite{DBLP:conf/acl/KnightCSRB18}.
% Specifically, 
% In all, we extract 13,568 examples in Story Commonsense that meet our requirements. The I/O are summarized in Table \ref{tab:task-definition}. 

Our goal is to first collect training and evaluation data for the proposed three tasks. 
Story Commonsense dataset manually annotates human motivations and emotions of the event in daily commonsense stories. 
It is an important resource for studying the causality of motivations, actions, and emotions in language-based individual activities. 
Note that, the actors of the actions are required to have both motivations and emotion labeling in our collected data \textbf{\textsc{Hail}}. 
% In order to obtain such \textit{(motivation, action, emotion)} samples, we utilize NLTK\footnote{http://www.nltk.org/} (a natural language processing toolkit) and design some rules. 
In order to obtain such \textit{(motivation, action, emotion)} samples, we align the motivation prediction and emotion prediction data of Story Commonsense by the story id and the character of the current story event. 
In all, we extract 13,568 examples in Story Commonsense that meet our requirements. Fig. \ref{fig:data_analysis} denotes the data statistics of label distributions in \textbf{\textsc{Hail}}, including motivations and emotions. The label distribution is relatively uniform, which is conducive to the learning of the model.

% \noindent\textbf{Data Analysis} We perform analysis about the gender bias of open-text actions in \textsc{Hail}. As is shown Fig. \ref{fig:se}, our dataset have a good distribution considering the gender of individual in all actions.

% This mechanism ensures that there is a clear and agreed-upon relationship between \textit{needs-action-emotion} in the story, and avoids subjectivity and ambiguity in SCT \cite{DBLP:conf/acl/SharmaABM18} and certain NLU tasks \cite{DBLP:conf/acl/NieWDBWK20}.

% \begin{table}[h]
%     \centering
%     \scalebox{0.8}{
%     \begin{tabular}{ll}
%     \toprule[1pt]
%     $\mathcal{H}$   &     Phillipa baked a cake. She put on chocolate icing. \\
%     $\mathcal{C}$   &     Phillipa\\
%     $\mathcal{M}$   &     love and belonging\\
%     $\mathcal{E}$   &     joy \\
%     $\mathcal{A}$   &     She gave it to her sister for her birthday.\\
%     \toprule[1pt]
%     \end{tabular}
%     }
%     \caption{Caption}
%     \label{tab:my_example}
% \end{table}

\section{Methodology}
\label{sec-5}

\subsection{Model for Emotion Understanding}
\label{sec-5-1}

Our method combines \textsc{Roberta}, Motivation Concept Knowledge Bases (MCKB) with a voting module component, which is shown in Fig. \ref{fig:model_cls}.
All input are refactored by prompt template and special tokens (Appendix \ref{appdix-prompt}) for improved understanding.

\subsubsection{Human Activity Encoder}
\label{sec-5-2}

Here, we use \textsc{Roberta} \cite{DBLP:journals/corr/abs-1907-11692} as our language-based human activity encoder.
It is a improved robust \textsc{Bert} \cite{DBLP:conf/naacl/DevlinCLT19} which shows state-of-the-art results in many NLP tasks. 
We use the hidden state representation of \texttt{<s>} as the sentence representation $h_s$.

\begin{figure*}[ht]
    \centering
    \includegraphics[width=1\textwidth]{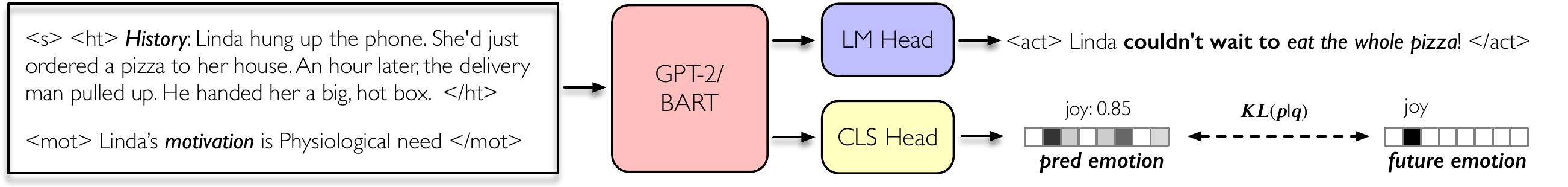}
    \caption{An example and the model architecture (GPT-2 or BART with language model head and emotion prediction head) for conditioned action generation task. Human mental states and key words in action are \textit{italic}. The words expressing future emotion in action are \textbf{bolden}.}
    \label{fig:model_gpt}
\end{figure*}

% \noindent\textbf{Need Concept KB:}
\subsubsection{Concept Knowledge Base}
For emotion understanding and motivation understanding tasks, we introduce knowledge bases to calculate the distribution of commonsense knowledge in language-based actions of all motivation/emotion categories.
In this paper, commonsense knowledge means commonsense concepts (i.e., words) with significant meanings that appear in language-based actions.

We build the Motivation Concept Knowledge Base (\textbf{MCKB}) in three steps.
Firstly, we extract representative commonsense concepts (details in Appendix \ref{appdix-kb}). 
Then, we count the number of occurrences of each commonsense concepts in the category of motivations.
% Next, the number of each concept's occurrences under each label is counted to form Matrix $M_{cn} \in \mathbb{R}^{d_c \times d_n}$, where $d_c$ and $d_n$ are separately the dimension of number of concepts and motivations.
The last step is to calculate word frequency.
The knowledge distribution of each concept is computed as below:
\begin{equation}
\mathcal{E}\left(c_{{j}}^{{s}_{{i}}}\right)=\frac{\text{Num}\left(c_{{j}}^{{s}_{{i}}}\right)}{\sum_{i=1}^{n} \text{Num}\left(c_{{j}}^{{s}_{{i}}}\right)} \times \frac{\mathcal{V}^{{s}_{{i}}}}{\mathcal{N}^{{s}_{{i}}}}
\label{p_important}
\end{equation}
where $c_{{j}}^{{s}_{{i}}}$ is $j$-th concept of $i$-th label ($s$ represents the current categorization), $\text{Num}$ is the count of concept $c_{{j}}^{{s}_{{i}}}$ occurrences. $\mathcal{V}$ is the vocab size. $\mathcal{N}$ is the total count of all concept occurrences.

\subsubsection{Classifier with Voting Gate}
\label{sec-5-2-2}
For emotion understanding task, we respectively calculate neural distribution of \textsc{Roberta} and knowledge distribution of MCKB. 
Lastly, we utilize a voting gate module to vote and integrate these two distributions.

\noindent\textbf{Neural Distribution of Encoder}
% \subsubsection{Neural Distribution of Encoder}
Once the sentence encoding $h_s$ is extracted, we then compute a probability distribution over labels, $P_z$, by the hidden representation from the classifier token $h_s\in \mathbb{R}^H$ through an MLP:
\begin{equation}
P_{z}=W_{2} \tanh \left(W_{1} h_{s}+b_{1}\right)
\label{p_roberta}
\end{equation}
where $W_{1} \in \mathbb{R}^{H \times H}$, $b_{1} \in \mathbb{R}^{H}$ and $W_{2} \in \mathbb{R}^{N \times H}$, $N$ is the number of labels. The model’s predicted answer corresponds to the label of motivations with the highest probability.

\noindent\textbf{Knowledge Distribution of KBs}
% \subsubsection{Prior Distribution of KBs}
First, we use NLP parsing methods to extract representative commonsense concepts corresponding to the current action. 
Second, we use each commonsense concept to retrieve the corresponding distribution in concept KBs.
In this way, the distribution of all commonsense knowledge $\{P_{c_1},P_{c_2},\dots,P_{c_n}\}$ in the current motivation category is obtained. 
More details, please refer to Appendix \ref{appdix-kb}.

\noindent\textbf{Voting Gate}
As shown in Fig. \ref{fig:model_cls}, the specific voting method is as follows:
\begin{equation}
P_{f} = \mathcal{F}_v(P_{z}, [P_{c_1},P_{c_2},...,P_{c_n}])
\label{p_voting}
\end{equation}
where $n$ is the number of related concepts to action. Among them, $\mathcal{F}_v$ denotes voting ensemble by pooling (such as \textsc{Aver}, \textsc{Max}, and \textsc{Sum} pooling), multilayer perceptron (MLP) or gating mechanisms. Finally, the selected label of the largest probability is used as the final prediction result.

% \subsection{Explainer}
% \label{sec-5-3}

% For emotion prediction, emotion abduction and action abduction tasks, the system required to generate a natural language explanation of the mental state according to the current task after obtaining the final prediction result. It helps researchers to better understand the relationship among motivations, actions and emotions. This paper uses the slot filling method, just fill the predicted result into the pre-set slot. The details are shown in Table \ref{app-3}. The output of the whole models will be a natural language explanation for emotions or actions.

\subsection{Model for Motivation Understanding}

Similar to the model for emotion understanding, we build emotion concept knowledge base (\textbf{ECKB}). 
The difference is the categorization and the dimension of \textbf{ECKB}. 
The remaining modules are the same as Emotion Understanding.

\subsection{Model for Conditioned Action Generation}
\label{sec-5-2-4}
As shown in Fig.~\ref{fig:model_gpt}, for conditioned action generation task, we employ pre-trained transformer~\cite{DBLP:conf/nips/VaswaniSPUJGKP17} based language models (LM) because of their exceptional performance across related NLG tasks ~\cite{DBLP:conf/emnlp/ForbesHSSC20,DBLP:conf/emnlp/RudingerSHBFBSC20,DBLP:conf/aaai/SakaguchiBBC20,ijcai2022wei}. Specifically, we select two standard text generation models for action prediction task:
\noindent\textbf{1. BART} \cite{DBLP:conf/acl/LewisLGGMLSZ20} is a encoder-decoder architecture;
\noindent\textbf{2. GPT2} \cite{Radford2019LanguageMA} is a single ``standard'' LM.
We call the models trained in our settings as \textbf{\textsc{Cog-Bart}} and \textbf{\textsc{Cog-Gpt2}} for performing experiments.
Besides, in order to control the direction of action generation is oriented to the future emotion, we adopt a emotion predictor to minimum the distance between emotion of action and the given emotion label.
To teach the model semantic information of the input text, which are motivations and all actions, we design prompt template for action generator (Table \ref{app-5} in Appendix).

\subsection{Training}
\label{sec-5-3}

\noindent \textbf{Emotion Understanding and Motivation Understanding}
% \subsubsection{Emotion Understanding and Motivation Understanding}
For encoders of these two tasks, we adopt the general MLP classification head and obtain the distribution on each label after fine-tuning with cross-entropy loss:
\begin{equation}
\mathcal{L}_{\text{CLS}}=-\sum_{i=1}^{n} p\left(x_{i}\right) \log \left(q\left(x_{i}\right)\right)
\label{loss_cls}
\end{equation}
where $x_i$ represent one sample in EU and MU tasks.

\noindent \textbf{Conditioned Action Generation}
% \subsubsection{Conditioned Action Generation}
\textsc{Gpt}-2 and \textsc{Bart} is trained to learn to produce the action $\mathcal{A}$ of the given history actions $\mathcal{H}$, motivation $\mathcal{M}$ and corresponding character $C$.
To achieve this goal, our approach is trained to maximize the conditional log-likelihood of predicting the object tokens of $\mathcal{A}$:
\begin{equation}
\mathcal{L}_{\text{LM}}=-\sum_{t=|\mathcal{H}|+|\mathcal{M}|+|\mathcal{C}|}^{|\mathcal{H}|+|\mathcal{M}|+|\mathcal{C}|+|\mathcal{A}|} \log P\left(x_{t} \mid x_{<t}\right)
\label{loss_gpt2_}
\end{equation}
What's more, the predicted emotion distribution of emotion predictor is supervised by the given emotion label distribution with KL-divergence.
\begin{equation}
\mathcal{L}_{\text{KL}}=\textbf{KL}(p(e_i)|q(e_i))
\label{loss_kl}
\end{equation}
where $p(e_i)$ is the predicted emotion distribution, and $q(e_i)$ is the given emotion label distribution.

To summarize, the total loss is:
\begin{equation}
\mathcal{L}=\lambda_1 \mathcal{L}_{\text{LM}} + \lambda_2 \mathcal{L}_{\text{KL}}
\label{loss_all}
\end{equation}
where $\lambda_*$ is the hyper-parameter controlling the proportion of its part.

\section{Experimental Setup}
\label{sec-6}

\subsection{Implement Details}
\label{sec-6-1}

We train baselines (\textsc{Gru} \cite{DBLP:journals/corr/ChungGCB14}, \textsc{Bert} \cite{DBLP:conf/naacl/DevlinCLT19} described in Appendix \ref{appdix-baseline}) and our models on 9k \textbf{\textsc{Hail}} training examples, then select hyper-parameters based on the best performing model on the dev set (2k), and then report results on the test set (2k).
We employ \textsc{Gpt2} large (1.5B), \textsc{Bart} large (680M) and \textsc{Roberta} large (340M) for our model. We implement our methods with HuggingFace\footnote{https://github.com/huggingface/transformers} \cite{DBLP:conf/emnlp/WolfDSCDMCRLFDS20} PyTorch \cite{DBLP:conf/nips/PaszkeGMLBCKLGA19}.
We use V-100 GPU to run the experiments.
More details, refer to Appendix \ref{appdix-imp}.

\subsection{Baselines}
\label{sec-6-x}

\noindent\textbf{1. GRU} \cite{DBLP:journals/corr/ChungGCB14} is a one-layer bi-GRU encodes the input text and concatenates the final time step hidden states from both directions to yield the sentence representation $h_s$.

\noindent\textbf{2. BERT} \cite{DBLP:conf/naacl/DevlinCLT19} is a standard pre-trained language model.
% To encode the input text using \textsc{BERT}, we follow the classification setup implementation described by authors.
We concatenate sentences using specific separator tokens (\texttt{[CLS]} and \texttt{[SEP]}). Finally, we take the hidden state representation of \texttt{[CLS]} in the last layer of BERT as the overall representation $h_s$ of sentence pairs;

\noindent\textbf{3. RoBERTa} \cite{DBLP:journals/corr/abs-1907-11692} is a improved robust BERT which shows state-of-the-art results in many NLP tasks. We use the hidden state representation of \texttt{<s>} as the sentence representation $h_s$.

\subsection{Metrics}
\label{sec-6-2}

\noindent \textbf{Automatic Metrics}
We report the micro-averaged precision (\textbf{P}), recall (\textbf{R}), and \textbf{F1} score\footnote{https://github.com/scikit-learn/scikit-learn} for emotion and motivation understanding tasks.

For conditioned action generation task, we adopt three automatic measures to evaluate the generated textual action distribution both on content quality and rationality. 
We use the following measures:
\textbf{(1)} Perplexity (\textbf{PPL}) as an indicator of fluency. A smaller value is better.
\textbf{(2) BLEU} \cite{DBLP:conf/acl/PapineniRWZ02} score with n is 1, 2, 4.
\textbf{(3) Rouge} \cite{DBLP:conf/naacl/LiGBGD16} score with n is 1, 2 or L.

\noindent \textbf{Human Evaluation Metrics}
We also conduct a human evaluation of generated action. Crowd-workers are required to evaluate actions on a 0-3 scale (3 being very good) from two different perspectives: (1) content quality to indicate whether the generated action is \textbf{fluent and coherent}, and (2) content rationality to assess whether it \textbf{follows the given motivations and emotions}.

\subsection{Automatic Evaluation}
\label{sec-6-3}

\noindent \textbf{Emotion Understanding}
% \subsection{Emotion Understanding}
We show results on the test set in Table \ref{table-cls-EP}. Our approach, which using prompt template, constructed KBs, \textsc{Roberta} and Voting module, achieves the highest score of all models. It is interesting that Emotion Prediction and Abduction are hard for pre-trained language models. In action abduction task, the results of all models are near to human performance. This further supports our hypothesis that emotion expectation plays a important role of motivation readiness for human taking actions.

\noindent \textbf{Motivation Understanding}
% \subsection{Motivation Understanding}
As shown in Table \ref{table-cls-MU}, we can conclude that our method outperform other models. However, the improvement in this task is small. It is possible that MU task needs the ability to reason. All in all, MU task is challengeable for the state-of-the-art models in natural language understanding tasks.

\noindent \textbf{Conditioned Action Generation}
% \subsection{Conditioned Action Generation}
Table \ref{table-gen} shows that our \textsc{Cog-Gpt2} and \textsc{Cog-Bart} outperforms all baselines, indicating that it can serve as good base action prediction model. We can conclude that the BLEU-1 score of \textsc{Cog-Gpt2} models is the best. 
For Rouge score, \textsc{Cog-Bart} model shows best performance.
One reason is that summarization task is helpful for generating text with larger recall.

\noindent \textbf{Summary}
In conclusion, our approach shows better performance than other implemented state-of-the-art models with the relationship of motivations, emotions and actions. The results of all tasks verify the feasibility of \method.

\begin{table}[t]
\centering
\scalebox{0.8}{
\begin{tabular}{lccc}
\toprule[1pt]
\bf Models  & \textbf{P}     & \textbf{R}     & \textbf{F1}                \\
\toprule[0.5pt]
\textsc{Gru} \cite{DBLP:journals/corr/ChungGCB14}   & 36.23 & 36.76 & 36.51           \\
\textsc{Bert}$_{\mbox{\scriptsize BASE}}$$^\dagger$ \cite{DBLP:conf/naacl/DevlinCLT19}   & 47.63 & 54.34 &  49.77      \\
\textsc{Bert}$_{\mbox{\scriptsize LARGE}}$$^\dagger$ \cite{DBLP:conf/naacl/DevlinCLT19}  & 53.95 & 55.23 &  53.23          \\
\textsc{Roberta}$_{\mbox{\scriptsize BASE}}$$^\dagger$ \cite{DBLP:journals/corr/abs-1907-11692}      & 51.47 & 55.64 & 53.09        \\
\textsc{Roberta}$_{\mbox{\scriptsize LARGE}}$$^\dagger$  \cite{DBLP:journals/corr/abs-1907-11692}    & 54.36 & 58.27 &  55.93     \\
\toprule[0.5pt]
\textsc{Ours}     & \bf 56.75 & \bf 60.39 & \bf 59.12    \\
\toprule[0.5pt]
(1) w/o \textsc{Roberta}    &    54.04     &     58.77    &  58.77  \\
(2) w/o Knowledge Base   & 54.73 & 58.90 &  57.62  \\
(3) w/o prompt template  & 54.95 & 58.23 &  56.23     \\
(4) w/o voting module  & 55.81 & 59.01 &  57.54     \\
\toprule[1pt]
\end{tabular}}
\caption{Results of Emotion Understanding. 
Our Approach is our proposed model. 
$^\dagger$ means following the experimental settings in papers.}
\label{table-cls-EP}
\end{table}

\begin{table}[t]
\centering
\scalebox{0.8}{
\begin{tabular}{lccc}
\toprule[1pt]
\bf Models & \bf P  & \bf R & \bf F1    \\
\toprule[0.5pt]
\textsc{Gru} \cite{DBLP:journals/corr/ChungGCB14}  & 40.53 & 40.27 & 40.89            \\
\textsc{Bert}$_{\mbox{\scriptsize BASE}}$$^\dagger$ \cite{DBLP:conf/naacl/DevlinCLT19}    & 60.57 & 60.80 & 60.28     \\
\textsc{Bert}$_{\mbox{\scriptsize LARGE}}$$^\dagger$ \cite{DBLP:conf/naacl/DevlinCLT19}   & 61.17 & 61.45 & 60.96         \\
\textsc{Roberta}$_{\mbox{\scriptsize BASE}}$$^\dagger$ \cite{DBLP:journals/corr/abs-1907-11692}        & 61.53 & 61.62 &  61.06        \\
\textsc{Roberta}$_{\mbox{\scriptsize LARGE}}$$^\dagger$  \cite{DBLP:journals/corr/abs-1907-11692}     & 63.50 & 63.97 &  63.57      \\
\toprule[0.5pt]
\textsc{Ours}      & \bf 64.57 & \bf 64.56 &  \bf 63.96   \\
\toprule[0.5pt]
(1) w/o \textsc{Roberta}   & 59.68 & 59.24 & 58.77  \\
(2) w/o Knowledge Base    & 62.34 & 62.35 &  62.57  \\
(3) w/o prompt template    & 63.22 & 63.76 &  62.19     \\
(4) w/o voting module  & 63.65 & 63.93 &  62.44    \\
\toprule[1pt]
\end{tabular}}
\caption{Results of Motivation Understanding. Our Approach is our proposed model. $^\dagger$ means following the experimental settings in papers.}
\label{table-cls-MU}
\end{table}

\begin{table*}[t]
\centering
\scalebox{0.8}{
\begin{tabular}{lccccccccc}
\toprule[1pt]
\multirow{2}{*}{\textbf{Models}} & \multicolumn{7}{c}{\bf Automatic Eval} & \multicolumn{2}{c}{\bf Human Eval} \\ \cline{2-10} 
 & PPL & BLEU-1 & BLEU-2 & BLEU-4 & Rouge-1 & Rouge-2 & Rouge-L & Content & Plausible \\ \hline
\textsc{Gpt2}+\textsc{Roc}$^\ddagger$ & 12.47 & 17.24 & 6.26 & 2.16 & 6.86 & 0.23 & 6.44 & 2.36 & 0.79 \\
\textsc{Gpt2}+\textsc{Hail}$^\ddagger$ & 11.83 & 16.46 & 5.81 & 1.92 & 7.32 & 0.38 & 6.71 & 2.24 & 0.56 \\ 
\midrule[0.5pt]
\textsc{Cog-Gpt2} & 6.85 & 22.48 & \bf 7.54 & \bf 2.85 & 10.86 & 1.05 & 10.25 & 2.79 & \bf 2.12 \\
\hdashline
w/o $\mathcal{E}$  & 7.99 & 22.29 & 7.58 & 2.81 & 10.22 & 0.94 & 9.66 & 2.85 & 1.79 \\
w/o $\mathcal{M}$ & 8.56 & 21.71 & 7.12 & 2.48 & 10.57 & 0.96 & 10.06 & 2.72 & 1.63 \\ 
\midrule[0.5pt]
\textsc{Cog-Bart} & \bf 6.58 & \bf 24.51 & 2.26 & 0.31 & \bf 18.71 & \bf 3.11 & \bf 17.24 & \bf 2.87 & 1.98 \\
\hdashline
w/o $\mathcal{E}$  & 7.65 & 23.62 & 2.01 & 0.22 & 17.68 & 2.74 & 16.25 & 2.86 & 1.58 \\
w/o $\mathcal{M}$ & 8.86 & 23.98 & 1.94 & 0.16 & 18.53 & 2.72 & 16.99 & 2.79 & 1.85 \\ 
\toprule[1pt]
\end{tabular}}
\caption{Automatic and human evaluation results of our \textsc{Cog-Gpt2} and \textsc{Cog-Bart} models on conditioned action generation. 
\textsc{Cog-Gpt2} and \textsc{Cog-Bart} are trained with the combination of character $\mathcal{C}$, motivation $\mathcal{M}$ and are required to predict emotion $\mathcal{E}$. 
$^\ddagger$ represents pre-training \textsc{Gpt2} with language model objective on the corresponding corpus. \textsc{Roc} is ROCStories \cite{DBLP:conf/naacl/MostafazadehCHP16} and \textbf{\textsc{Hail}} is the train set of our proposed dataset. }
\label{table-gen}
\end{table*}

\subsection{Human Evaluation}
\label{sec-5-4}

We also performed manual evaluation for conditioned action prediction. 
We randomly selected 100 instances from the test set and used the evaluated model to generate actions. 
In our work, we compare the generated stories in pairs, and each pair is evaluated by 3 judges. 
The last two columns of Table \ref{table-gen} report the average improvements as well as absolute scores for content quality and rationality. 
We can conclude that models with our designed prompt template and training loss outperform the models pre-trained on story corpus with the language model objective. 
It is interesting that the content of generated actions is fluent and grammatical, which indicates that \textsc{Gpt2} and \textsc{Bart} is good at organize natural language.

\section{Analysis and Discussion}
\label{sec-8}

\begin{table}[t] % Human A/B Test
\centering
\scalebox{0.7}{
	\begin{tabular}{lcccc}
		\toprule[1pt]
		\bf Methods  & \textbf{Win}  & \textbf{Loss}   & \textbf{Tie} & $\kappa$   \\
		\midrule
		\textsc{Cog-Gpt2} v.s. \textsc{Gpt2}+\textsc{Roc}    &  \bf 54.2\%    &   19.5\%   &  26.3\%   &  30.8  \\
		\textsc{Cog-Gpt2} v.s. \textsc{Gpt2}+\textsc{Hail}     &  \bf 49.4\%    &   18.7\%   &   31.9\%   &  29.6   \\
		\textsc{Cog-Bart} v.s. \textsc{Gpt2}+\textsc{Roc}    &  \bf 53.3\%    &   18.4\%   &   28.3\%    &  28.9   \\
		\textsc{Cog-Bart} v.s. \textsc{Gpt2}+\textsc{Hail}   &  \bf 54.3\%    &   14.6\%   &   31.1\%    &  31.3 \\
		\bottomrule[1pt]
    \end{tabular}}
\caption{Human A/B Test of \method. Results show that \method performs baseline models sufficiently. 
$\kappa$ denotes Fleiss’ kappa (all are fair agreement or moderate agreement).
The p-value of scores < 0.05 in sign test.}
\label{table-abtest}
\end{table}

\subsection{Ablation Study}
\label{sec-8-1}

To analyze the importance of different modules in our baseline models, we perform ablation study on our approach in emotion and motivation understanding.
As shown in Table \ref{table-cls-EP} and Table \ref{table-cls-MU}, (1) denotes that the semantic representation of \textsc{Roberta} is crucial for understanding tasks.
Compared (1) and (2), we find that \textsc{Roberta} and KBs have the similar scores in motivation understanding.
(3) and (4) indicate the importance of prompt template and voting modules designed for \textsc{Roberta}.

\begin{figure}[t]
    \centering
    \includegraphics[width=0.48\textwidth]{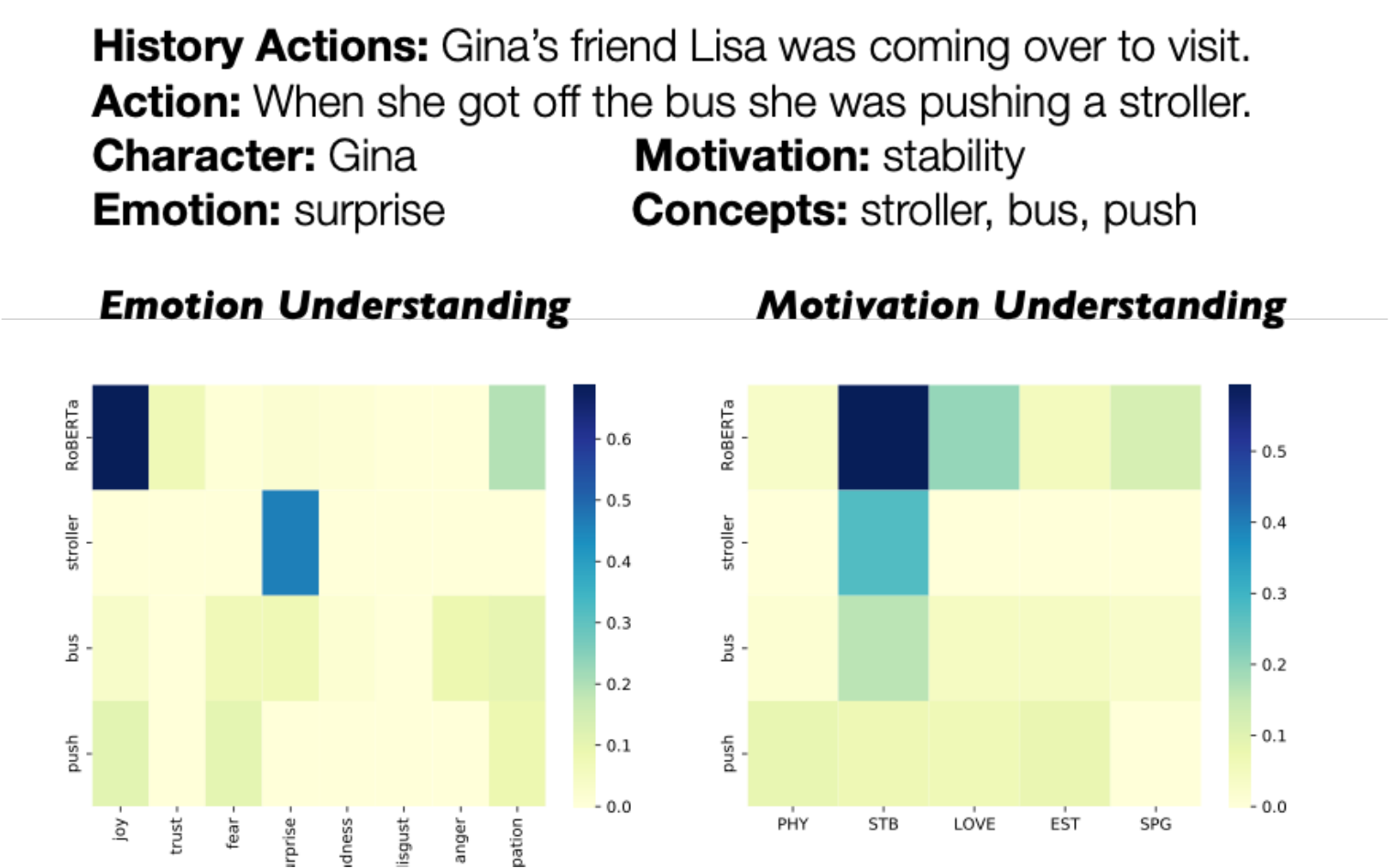}
    \caption{Case study for emotion and motivation understanding. Our framework can give better interpretability.}
    \label{fig:case-cls}
\end{figure}

\subsection{Human A/B Test}
\label{sec-8-2}
Human A/B test is also conducted. 
We try to directly compare our model with other baselines. 
We randomly sample 100 examples each for our model and baseline models. 
Three annotators are given generated responses from either our model or baselines in random order and are asked to choose a better one. 
They can either choose one of the responses or select “Tie” when the quality of provided options are hard to access.
Results in Table \ref{table-abtest} confirm that the responses from \method are more preferred by human judges.

\subsection{Case Study}
\label{sec-8-3}
\noindent \textbf{Emotion and Motivation Understanding} Fig. \ref{fig:case-cls} illustrates the distribution of \textsc{Roberta} and our concept knowledge base. Our method can bring with better interpretability with the knowledge of key words in human activities. Fig. \ref{fig:case-cls} shows that knowledge base predicts correctly in emotion understanding and help motivation understanding.

\noindent \textbf{Conditioned Action Generation} Since the proposed models can generate actions conditioned on one's motivation, they can be used to unfold action in diverse situations for a combination of history actions, character, motivation, and emotion.
We demonstrate this capability in Table \ref{table-case}. $\surd$ means reasonable. \includegraphics[width=0.8em]{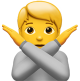} means that the generated action can not express the corresponding aspect. \includegraphics[width=0.8em]{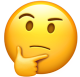} represents that the consistency is debatable. 
% Fig. \ref{table-case} shows eight generated actions that conditioned on the same motivation.
It can be concluded that \textbf{motivation and emotion are all important for action generation}.
\textsc{Cog-Gpt}-2 tends to generate short but reasonable actions. 
But actions generated by \textsc{Cog-Bart} usually are long but repetitive.
From the samples tagged by thinking face, we can see that only motivations or emotions are hard to make action prediction.
% Interestingly, \textsc{Gpt}-2 pre-trained on ROC Stories or \textbf{\textsc{Hail}} could learn the personality or other feature of a specific individual.
% We provide more cases with different input in Appendix \ref{appdix-case}.

\begin{figure*}[ht]
    \centering
    \includegraphics[width=0.9\textwidth]{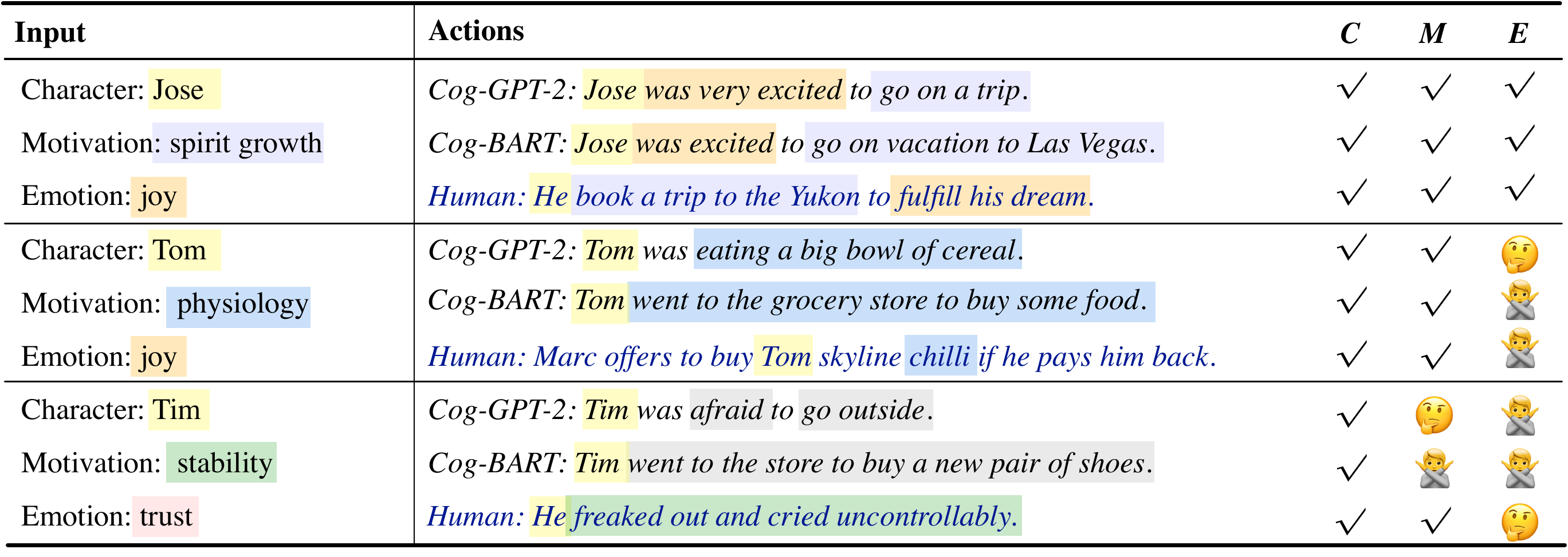}
    \caption{Case study with different inputs on action generation task. The actions are generated by \textsc{Cog-Gpt}-2, \textsc{Cog-Bart} or human writing with the specific character, motivation and emotion. $\surd$ means reasonable. \includegraphics[width=0.8em]{fig/vote_no.pdf} means that the generated action can not express the corresponding aspect. \includegraphics[width=0.8em]{fig/thinking.pdf} represents that the consistency is debatable. 
The colored text means generated action satisfy the aspects of mental states.}
    \label{fig:newcase}
\end{figure*}

In order to analyze the effectiveness of our baseline models for action generation, we also perform some case studies with different inputs. The actions, with the inputs of specific history actions $\mathcal{H}$, character $\mathcal{C}$, motivation $\mathcal{M}$ or emotion $\mathcal{E}$, are generated by \textsc{Cog-Gpt}-2, \textsc{Cog-Bart} or human writing. $\surd$ means consistent to the input, while $\times$ means that the generated action can not express the corresponding aspect. The thinking face shows that the consistency of input and generated action is debatable. From the first line of Fig. \ref{fig:newcase}, \textit{Jose has a spirit growth need} and \textit{Jose's emotional expectation is joy}. As demonstrated by the three actions, we can conclude that the \textsc{Cog-Gpt}-2, \textsc{Cog-Bart} based models can generate reasonable actions. Interestingly, \textsc{Cog-Bart} can guess that the destination of the trip is Las Vegas, which is competitive to human writing. In the second example, all actions can not clearly express that the emotional expectation is \textit{joy}, where \textit{a big bowl} generated by \textsc{Cog-Gpt}-2 could somewhat show the happiness of Tom. Last but not least, \textit{Tim was afraid to go outside} and \textit{Tim went to the store to buy a new pair of shoes} are plausible corresponding to the stability need. It is possible that stability need is more abstract for pre-trained language model (PLM) to understand. Besides, we can find that all actions lack the expression of \textit{trust} emotion. One reason is that the PLM based models are insensitive to emotional inputs, which is challengeable in future work.

% \colorlet{soulyellow}{yellow!50}
% \colorlet{soullime}{lime!50}
% \colorlet{soulblue}{blue!20}
\begin{table}[t]
\centering
\scalebox{0.7}{
\begin{tabular}{llc}
\toprule[1pt]
\multirow{3}{*}{\textbf{Models}}   & \multicolumn{1}{p{5.5cm}}{\textbf{Given} $\mathcal{H}$ = \textit{Kim and her glass went on a field trip to an aquarium.} $\mathcal{C}$=\sethlcolor{soulblue}\hl{Kim}, $\mathcal{M}$=\sethlcolor{soulyellow}\hl{spirit growth}, $\mathcal{E}$=\sethlcolor{soullime}\hl{joy}}  & \multirow{3}{*}{\textbf{R}}   \\
\toprule[0.5pt]
\textsc{Gpt2}+\textsc{Roc}                &   \textit{\sethlcolor{soulblue}\hl{Kim} has a very good sense of \colorbox{soullime}{humor}.}       &   \includegraphics[width=0.8em]{fig/vote_no.pdf} \\
\textsc{Gpt2}+\textsc{HAIL}                &   \textit{The personality of \sethlcolor{soulblue}\hl{Kim} could be \sethlcolor{soullime}\hl{humorous}.}     & \includegraphics[width=0.8em]{fig/vote_no.pdf} \\
\toprule[0.5pt]
\textsc{Cog-Gpt2}      &   \textit{\sethlcolor{soulblue}\hl{She} \sethlcolor{soullime}\hl{was able to} \sethlcolor{soulyellow}\hl{get a job at a local restaurant}.}     & $\surd$  \\
\hdashline
w/o $\mathcal{E}$     &   \textit{\sethlcolor{soulblue}\hl{She} was \sethlcolor{soulyellow}\hl{so excited to go}.}     &  \includegraphics[width=0.8em]{fig/thinking.pdf} \\
w/o $\mathcal{M}$      &   \textit{\sethlcolor{soulblue}\hl{Kim} was a very \sethlcolor{soulyellow}\hl{hard working} woman.}     &  \includegraphics[width=0.8em]{fig/vote_no.pdf} \\
\toprule[0.5pt]
\textsc{Cog-Bart}                &   \textit{\sethlcolor{soulblue}\hl{Kim} \sethlcolor{soullime}\hl{had always} \sethlcolor{soulyellow}\hl{wanted to go to the beach}.}     & $\surd$  \\
\hdashline
w/o $\mathcal{E}$            &   \textit{\sethlcolor{soulblue}\hl{Kim} \sethlcolor{soullime}\hl{had always} \sethlcolor{soulyellow}\hl{wanted to be a pilot}.}     & $\surd$   \\
w/o $\mathcal{M}$            &   \textit{\sethlcolor{soulblue}\hl{Kim} and her friends \sethlcolor{soulyellow}\hl{decided to go on a date}.}      &    \includegraphics[width=0.8em]{fig/thinking.pdf}  \\
\toprule[0.5pt]
Human      &   \textit{\sethlcolor{soulblue}\hl{Kim} \sethlcolor{soulyellow}\hl{enjoyed} \sethlcolor{soullime}\hl{looking at the sea creatures}.}     & $\surd$   \\
\toprule[1pt]
\end{tabular}}
\caption{Case study of conditioned action generation task for all models, which are tested with history actions, character \sethlcolor{soulblue}\hl{Kim}, motivation \sethlcolor{soulyellow}\hl{spirit growth}, emotion \sethlcolor{soullime}\hl{joy}. 
Rationality is abbreviated as \textbf{R}.
% The last line is the golden human-written action. 
% $\surd$ means reasonable. \includegraphics[width=0.8em]{fig/vote_no.pdf} means that the generated action can not express the corresponding aspect. \includegraphics[width=0.8em]{fig/thinking.pdf} represents that the consistency is debatable. 
The colored text means generated action satisfy the aspects of mental states.}
\label{table-case}
\end{table}

\subsection{Visualization Analysis}
\label{sec-8-4}

\begin{figure}[t]
    \centering
    \includegraphics[width=0.48\textwidth]{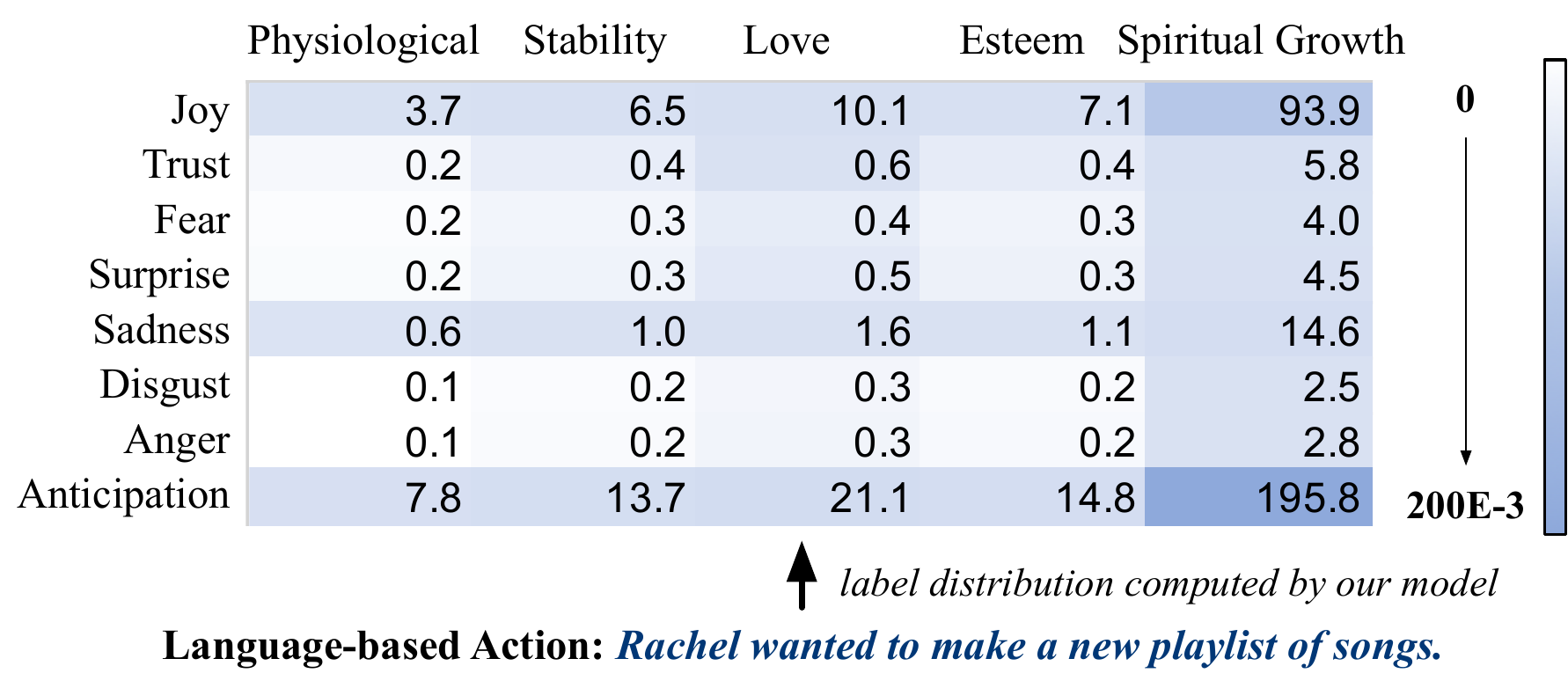}
    \caption{Visualization of final distributions of model. Each element of the above matrix uses scientific notation, and the exponent is $-3$.}
    \label{fig:vis}
\end{figure}

To verify the claim \method \textit{can reveal the essential relationship among motivations, emotions, and actions}, we conduct a visualization analysis of relationships among motivations, actions, and emotions with our approach.
Fig. \ref{fig:vis} demonstrates the matrix of final prediction probability of motivations and emotions in emotion understanding tasks. 
The matrix makes motivations, actions and emotions close together and shows that motivations (spiritual growth) have the future emotion (i.e. anticipation($195.8E^{-3}$)).
With this matrix, we can better reveal the essential relationship among motivations, emotions, and actions.
Therefore, we can supply more deep explanations about the relationship of motivations, actions, and emotions based on visualization analysis.

\section{Related Work}
\label{sec-2}

There have been many large-scale language-based resources to explore human mental state, such as motivations and emotions.
% Weak supervision and reinforcement learning were used in \cite{DBLP:conf/naacl/DingR18} to optimize the abstract concepts in the description of human motivations to assist in the classification of events.
\cite{DBLP:conf/acl/SmithCSRA18,DBLP:conf/aaai/SapBABLRRSC19,DBLP:journals/corr/abs-2010-05953} explore the human mental states in narrative text with series of ``if-then'' relationships.
SOCIAL IQA was introduced by \cite{DBLP:journals/corr/abs-1904-09728} for probing emotional and social intelligence in a variety of everyday situations.
% Each event in Event2mind involved one or two participants, and three tasks are proposed to predict the intentions and reactions of the main participants and the reactions of other participants.
Most similar to our work, \citet{DBLP:conf/acl/KnightCSRB18} put forward Story Commonsense, which is the causal reason for the changes in the psychological state of the characters in the story.

Recently, a lot of work has begun to consider introducing various mental state of human beings into sentiment analysis and other NLP downstream tasks.
\citet{DBLP:journals/corr/LiH15b} explore the importance of human motivations for sentiment analysis and consider emotion as a specific event or entity that realizes the mental state of human satisfaction with oneself. 
\citet{DBLP:conf/acl/OtaniH19} regard human motivation as the driving force of human emotions, and take motivation detection as the first step of emotion detection, which improves the sentiment analysis of evaluation.
\cite{DBLP:conf/emnlp/DuDLL19,Ammanabrolu2021AutomatedSV,DBLP:conf/emnlp/XuPSPFAC20,DBLP:conf/emnlp/BrahmanC20} use the knowledge generated by COMET regarded the psychological state of COMET as a condition for story generation.

All in all, the existing language-based resources and works focus on the binary relationship between action and each mental state.
Diversely, we first propose a cognitive framework that aims to analyze comprehensive relationships among motivations, emotions and actions in language-based human individual activities.

\section{Conclusion and Future work}
\label{sec-10}
In this paper, we propose a \textbf{Co}gnitive Fra\textbf{m}ework of Hu\textbf{m}an \textbf{A}ctivities (\method). 
To verify the effectiveness of our cognitive framework, we introduce three challenging NLP tasks, automatically construct a dataset \textbf{\textsc{Hail}}, and propose the corresponding methods. 
Experimental results show a better understanding of the relationship among motivations, emotions and actions under our \method than existing methods.

Modeling the relationship among motivations, emotions and actions in human activities can allow researchers to reason the essential causes of human activities from the cognitive perspective and supply reasonable explanations. In future work, we will explore \method on various NLP downstream applications, such as intelligent dialogue, controllable text generation and public opinion analysis.

\section*{Acknowledgement}

We thank all anonymous reviewers for their constructive comments and useful advice. 
Also thanks for the discussion with Yunpeng Li, Yajing Sun, Yongxiu Xu, Ping Guo, Xinyu Zhang, Yao Dong and Yige Chen. 
This work is supported by the National Natural Science Foundation of China (No.62006222 and No.U21B2009).
Thanks for organizers and the proposed pre-trained language models, data, codes. 

\textbf{Contribution List} Yuqiang Xie: Idea, Paper Writing, Coding; Yue Hu: Guiding, Discussion; Wei Peng: Discussion, Coding; Guanqun Bi: Discussion; Luxi Xing: Review.

Thanks for the hard work and dedication of all team members.

\bibliography{anthology,custom}

\begin{thebibliography}{43}
\expandafter\ifx\csname natexlab\endcsname\relax\def\natexlab#1{#1}\fi

\bibitem[{Ammanabrolu et~al.(2021)Ammanabrolu, Cheung, Broniec, and
  Riedl}]{Ammanabrolu2021AutomatedSV}
Prithviraj Ammanabrolu, W.~Cheung, William Broniec, and Mark~O. Riedl. 2021.
\newblock Automated storytelling via causal, commonsense plot ordering.
\newblock In \emph{AAAI}.

\bibitem[{Brahman and Chaturvedi(2020)}]{DBLP:conf/emnlp/BrahmanC20}
Faeze Brahman and Snigdha Chaturvedi. 2020.
\newblock \href {https://doi.org/10.18653/v1/2020.emnlp-main.426} {Modeling
  protagonist emotions for emotion-aware storytelling}.
\newblock In \emph{Proceedings of the 2020 Conference on Empirical Methods in
  Natural Language Processing, {EMNLP} 2020, Online, November 16-20, 2020},
  pages 5277--5294. Association for Computational Linguistics.

\bibitem[{Cabanac(2002)}]{Cabanac2002WhatIE}
Michel Cabanac. 2002.
\newblock What is emotion?
\newblock \emph{Behavioural Processes}, 60:69--83.

\bibitem[{Cacioppo and Gardner(1999)}]{emotion-annurev}
John~T. Cacioppo and Wendi~L. Gardner. 1999.
\newblock \href {https://doi.org/10.1146/annurev.psych.50.1.191} {Emotion}.
\newblock \emph{Annual Review of Psychology}, 50:191--214.
\newblock PMID: 10074678.

\bibitem[{Chung et~al.(2014)Chung, G{\"{u}}l{\c{c}}ehre, Cho, and
  Bengio}]{DBLP:journals/corr/ChungGCB14}
Junyoung Chung, {\c{C}}aglar G{\"{u}}l{\c{c}}ehre, KyungHyun Cho, and Yoshua
  Bengio. 2014.
\newblock \href {http://arxiv.org/abs/1412.3555} {Empirical evaluation of gated
  recurrent neural networks on sequence modeling}.
\newblock \emph{CoRR}, abs/1412.3555.

\bibitem[{Devlin et~al.(2019)Devlin, Chang, Lee, and
  Toutanova}]{DBLP:conf/naacl/DevlinCLT19}
Jacob Devlin, Ming{-}Wei Chang, Kenton Lee, and Kristina Toutanova. 2019.
\newblock \href {https://doi.org/10.18653/v1/n19-1423} {{BERT:} pre-training of
  deep bidirectional transformers for language understanding}.
\newblock In \emph{Proceedings of the 2019 Conference of the North American
  Chapter of the Association for Computational Linguistics: Human Language
  Technologies, {NAACL-HLT} 2019, Minneapolis, MN, USA, June 2-7, 2019, Volume
  1 (Long and Short Papers)}, pages 4171--4186. Association for Computational
  Linguistics.

\bibitem[{Du et~al.(2019)Du, Ding, Liu, and Li}]{DBLP:conf/emnlp/DuDLL19}
Li~Du, Xiao Ding, Ting Liu, and Zhongyang Li. 2019.
\newblock \href {https://doi.org/10.18653/v1/D19-1270} {Modeling event
  background for if-then commonsense reasoning using context-aware variational
  autoencoder}.
\newblock In \emph{Proceedings of the 2019 Conference on Empirical Methods in
  Natural Language Processing and the 9th International Joint Conference on
  Natural Language Processing, {EMNLP-IJCNLP} 2019, Hong Kong, China, November
  3-7, 2019}, pages 2682--2691. Association for Computational Linguistics.

\bibitem[{Forbes et~al.(2020)Forbes, Hwang, Shwartz, Sap, and
  Choi}]{DBLP:conf/emnlp/ForbesHSSC20}
Maxwell Forbes, Jena~D. Hwang, Vered Shwartz, Maarten Sap, and Yejin Choi.
  2020.
\newblock \href {https://doi.org/10.18653/v1/2020.emnlp-main.48} {Social
  chemistry 101: Learning to reason about social and moral norms}.
\newblock In \emph{Proceedings of the 2020 Conference on Empirical Methods in
  Natural Language Processing, {EMNLP} 2020, Online, November 16-20, 2020},
  pages 653--670. Association for Computational Linguistics.

\bibitem[{Hamilton et~al.(2016)Hamilton, Clark, Leskovec, and
  Jurafsky}]{DBLP:conf/emnlp/HamiltonCLJ16}
William~L. Hamilton, Kevin Clark, Jure Leskovec, and Dan Jurafsky. 2016.
\newblock \href {https://doi.org/10.18653/v1/d16-1057} {Inducing
  domain-specific sentiment lexicons from unlabeled corpora}.
\newblock In \emph{Proceedings of the 2016 Conference on Empirical Methods in
  Natural Language Processing, {EMNLP} 2016, Austin, Texas, USA, November 1-4,
  2016}, pages 595--605. The Association for Computational Linguistics.

\bibitem[{Hull(1974)}]{Hull1974EssentialsOB}
C.~L. Hull. 1974.
\newblock Essentials of behavior.
\newblock In \emph{New Haven: Published for the Institute of Human Relations by
  Yale University Press}.

\bibitem[{Hwang et~al.(2020)Hwang, Bhagavatula, Bras, Da, Sakaguchi, Bosselut,
  and Choi}]{DBLP:journals/corr/abs-2010-05953}
Jena~D. Hwang, Chandra Bhagavatula, Ronan~Le Bras, Jeff Da, Keisuke Sakaguchi,
  Antoine Bosselut, and Yejin Choi. 2020.
\newblock \href {http://arxiv.org/abs/2010.05953} {{COMET-ATOMIC} 2020: On
  symbolic and neural commonsense knowledge graphs}.
\newblock \emph{CoRR}, abs/2010.05953.

\bibitem[{Kagan(2007)}]{Kagan2007WhatIE}
Jerome Kagan. 2007.
\newblock What is emotion?: History, measures, and meanings.
\newblock In \emph{Yale University Press}.

\bibitem[{Lewis et~al.(2020)Lewis, Liu, Goyal, Ghazvininejad, Mohamed, Levy,
  Stoyanov, and Zettlemoyer}]{DBLP:conf/acl/LewisLGGMLSZ20}
Mike Lewis, Yinhan Liu, Naman Goyal, Marjan Ghazvininejad, Abdelrahman Mohamed,
  Omer Levy, Veselin Stoyanov, and Luke Zettlemoyer. 2020.
\newblock \href {https://doi.org/10.18653/v1/2020.acl-main.703} {{BART:}
  denoising sequence-to-sequence pre-training for natural language generation,
  translation, and comprehension}.
\newblock In \emph{Proceedings of the 58th Annual Meeting of the Association
  for Computational Linguistics, {ACL} 2020, Online, July 5-10, 2020}, pages
  7871--7880. Association for Computational Linguistics.

\bibitem[{Li et~al.(2016)Li, Galley, Brockett, Gao, and
  Dolan}]{DBLP:conf/naacl/LiGBGD16}
Jiwei Li, Michel Galley, Chris Brockett, Jianfeng Gao, and Bill Dolan. 2016.
\newblock \href {https://doi.org/10.18653/v1/n16-1014} {A diversity-promoting
  objective function for neural conversation models}.
\newblock In \emph{{NAACL} {HLT} 2016, The 2016 Conference of the North
  American Chapter of the Association for Computational Linguistics: Human
  Language Technologies, San Diego California, USA, June 12-17, 2016}, pages
  110--119. The Association for Computational Linguistics.

\bibitem[{Li and Hovy(2017)}]{DBLP:journals/corr/LiH15b}
Jiwei Li and Eduard Hovy. 2017.
\newblock \href {https://doi.org/10.1007/978-3-319-55394-8_3}
  {\emph{Reflections on Sentiment/Opinion Analysis}}, pages 41--59. Springer
  International Publishing, Cham.

\bibitem[{Liu et~al.(2019)Liu, Ott, Goyal, Du, Joshi, Chen, Levy, Lewis,
  Zettlemoyer, and Stoyanov}]{DBLP:journals/corr/abs-1907-11692}
Yinhan Liu, Myle Ott, Naman Goyal, Jingfei Du, Mandar Joshi, Danqi Chen, Omer
  Levy, Mike Lewis, Luke Zettlemoyer, and Veselin Stoyanov. 2019.
\newblock \href {http://arxiv.org/abs/1907.11692} {Roberta: {A} robustly
  optimized {BERT} pretraining approach}.
\newblock \emph{CoRR}, abs/1907.11692.

\bibitem[{Maslow(1943)}]{Maslow2013ATO}
Abraham~Harold Maslow. 1943.
\newblock \href
  {https://pdfs.semanticscholar.org/799e/de7676bbb166a33d61ae108436c4eb08c419.pdf?_ga=2.194837167.383461422.1605675815-1123820971.1540522923}
  {A theory of human motivation}.
\newblock In \emph{Psychological review}.

\bibitem[{Mohammad and Turney(2013)}]{DBLP:journals/ci/MohammadT13}
Saif Mohammad and Peter~D. Turney. 2013.
\newblock \href {https://doi.org/10.1111/j.1467-8640.2012.00460.x}
  {Crowdsourcing a word-emotion association lexicon}.
\newblock \emph{Comput. Intell.}, 29(3):436--465.

\bibitem[{Mostafazadeh et~al.(2016)Mostafazadeh, Chambers, He, Parikh, Batra,
  Vanderwende, Kohli, and Allen}]{DBLP:conf/naacl/MostafazadehCHP16}
Nasrin Mostafazadeh, Nathanael Chambers, Xiaodong He, Devi Parikh, Dhruv Batra,
  Lucy Vanderwende, Pushmeet Kohli, and James~F. Allen. 2016.
\newblock \href {https://doi.org/10.18653/v1/n16-1098} {A corpus and cloze
  evaluation for deeper understanding of commonsense stories}.
\newblock In \emph{{NAACL} {HLT} 2016, The 2016 Conference of the North
  American Chapter of the Association for Computational Linguistics: Human
  Language Technologies, San Diego California, USA, June 12-17, 2016}, pages
  839--849. The Association for Computational Linguistics.

\bibitem[{Nie et~al.(2020)Nie, Williams, Dinan, Bansal, Weston, and
  Kiela}]{DBLP:conf/acl/NieWDBWK20}
Yixin Nie, Adina Williams, Emily Dinan, Mohit Bansal, Jason Weston, and Douwe
  Kiela. 2020.
\newblock \href {https://doi.org/10.18653/v1/2020.acl-main.441} {Adversarial
  {NLI:} {A} new benchmark for natural language understanding}.
\newblock In \emph{Proceedings of the 58th Annual Meeting of the Association
  for Computational Linguistics, {ACL} 2020, Online, July 5-10, 2020}, pages
  4885--4901. Association for Computational Linguistics.

\bibitem[{Otani and Hovy(2019)}]{DBLP:conf/acl/OtaniH19}
Naoki Otani and Eduard~H. Hovy. 2019.
\newblock \href {https://doi.org/10.18653/v1/p19-1461} {Toward comprehensive
  understanding of a sentiment based on human motives}.
\newblock In \emph{Proceedings of the 57th Conference of the Association for
  Computational Linguistics, {ACL} 2019, Florence, Italy, July 28- August 2,
  2019, Volume 1: Long Papers}, pages 4672--4677. Association for Computational
  Linguistics.

\bibitem[{Panksepp(1998)}]{Panksepp1998AffectiveNT}
Jaak Panksepp. 1998.
\newblock Affective neuroscience: The foundations of human and animal emotions.
\newblock \emph{Psychology}.

\bibitem[{Papineni et~al.(2002)Papineni, Roukos, Ward, and
  Zhu}]{DBLP:conf/acl/PapineniRWZ02}
Kishore Papineni, Salim Roukos, Todd Ward, and Wei{-}Jing Zhu. 2002.
\newblock \href {https://doi.org/10.3115/1073083.1073135} {Bleu: a method for
  automatic evaluation of machine translation}.
\newblock In \emph{Proceedings of the 40th Annual Meeting of the Association
  for Computational Linguistics, July 6-12, 2002, Philadelphia, PA, {USA}},
  pages 311--318. {ACL}.

\bibitem[{Paszke et~al.(2019)Paszke, Gross, Massa, Lerer, Bradbury, Chanan,
  Killeen, Lin, Gimelshein, Antiga, Desmaison, K{\"{o}}pf, Yang, DeVito,
  Raison, Tejani, Chilamkurthy, Steiner, Fang, Bai, and
  Chintala}]{DBLP:conf/nips/PaszkeGMLBCKLGA19}
Adam Paszke, Sam Gross, Francisco Massa, Adam Lerer, James Bradbury, Gregory
  Chanan, Trevor Killeen, Zeming Lin, Natalia Gimelshein, Luca Antiga, Alban
  Desmaison, Andreas K{\"{o}}pf, Edward Yang, Zachary DeVito, Martin Raison,
  Alykhan Tejani, Sasank Chilamkurthy, Benoit Steiner, Lu~Fang, Junjie Bai, and
  Soumith Chintala. 2019.
\newblock \href
  {http://papers.nips.cc/paper/9015-pytorch-an-imperative-style-high-performance-deep-learning-library}
  {Pytorch: An imperative style, high-performance deep learning library}.
\newblock In \emph{Advances in Neural Information Processing Systems 32: Annual
  Conference on Neural Information Processing Systems 2019, NeurIPS 2019,
  December 8-14, 2019, Vancouver, BC, Canada}, pages 8024--8035.

\bibitem[{Peng et~al.(2022{\natexlab{a}})Peng, Hu, Xie, Xing, and
  Sun}]{Wei2022CogIntAc}
Wei Peng, Yue Hu, Yuqiang Xie, Luxi Xing, and Yajing Sun. 2022{\natexlab{a}}.
\newblock \href {https://doi.org/10.48550/arXiv.2205.03540} {Cogintac: Modeling
  the relationships between intention, emotion and action in interactive
  process from cognitive perspective}.
\newblock \emph{CoRR}, abs/2205.03540.

\bibitem[{Peng et~al.(2022{\natexlab{b}})Peng, Hu, Xing, Xie, Sun, and
  Li}]{ijcai2022wei}
Wei Peng, Yue Hu, Luxi Xing, Yuqiang Xie, Yajing Sun, and Yunpeng Li.
  2022{\natexlab{b}}.
\newblock \href {https://doi.org/10.24963/ijcai.2022/600} {Control globally,
  understand locally: {A} global-to-local hierarchical graph network for
  emotional support conversation}.
\newblock In \emph{Proceedings of the Thirty-First International Joint
  Conference on Artificial Intelligence, {IJCAI} 2022, Vienna, Austria, 23-29
  July 2022}, pages 4324--4330. ijcai.org.

\bibitem[{Plutchik(1980)}]{Plutchik1980AGP}
Robert Plutchik. 1980.
\newblock \href {https://doi.org/10.1016/B978-0-12-558701-3.50007-7} {A general
  psychoevolutionary theory of emotion}.
\newblock In \emph{Theories of emotion}.

\bibitem[{Pontiki et~al.(2016)Pontiki, Galanis, Papageorgiou, Androutsopoulos,
  Manandhar, Al{-}Smadi, Al{-}Ayyoub, Zhao, Qin, Clercq, Hoste, Apidianaki,
  Tannier, Loukachevitch, Kotelnikov, Bel, Zafra, and
  Eryigit}]{DBLP:conf/semeval/PontikiGPAMAAZQ16}
Maria Pontiki, Dimitris Galanis, Haris Papageorgiou, Ion Androutsopoulos,
  Suresh Manandhar, Mohammad Al{-}Smadi, Mahmoud Al{-}Ayyoub, Yanyan Zhao, Bing
  Qin, Orph{\'{e}}e~De Clercq, V{\'{e}}ronique Hoste, Marianna Apidianaki,
  Xavier Tannier, Natalia~V. Loukachevitch, Evgeniy~V. Kotelnikov, N{\'{u}}ria
  Bel, Salud Mar{\'{\i}}a~Jim{\'{e}}nez Zafra, and G{\"{u}}lsen Eryigit. 2016.
\newblock \href {https://doi.org/10.18653/v1/s16-1002} {Semeval-2016 task 5:
  Aspect based sentiment analysis}.
\newblock In \emph{Proceedings of the 10th International Workshop on Semantic
  Evaluation, SemEval@NAACL-HLT 2016, San Diego, CA, USA, June 16-17, 2016},
  pages 19--30. The Association for Computer Linguistics.

\bibitem[{Radford et~al.(2019)Radford, Wu, Child, Luan, Amodei, and
  Sutskever}]{Radford2019LanguageMA}
Alec Radford, Jeff Wu, Rewon Child, David Luan, Dario Amodei, and Ilya
  Sutskever. 2019.
\newblock \href
  {https://d4mucfpksywv.cloudfront.net/better-language-models/language-models.pdf}
  {Language models are unsupervised multitask learners}.
\newblock In \emph{OpenAI Blog}.

\bibitem[{Rashkin et~al.(2018{\natexlab{a}})Rashkin, Bosselut, Sap, Knight, and
  Choi}]{DBLP:conf/acl/KnightCSRB18}
Hannah Rashkin, Antoine Bosselut, Maarten Sap, Kevin Knight, and Yejin Choi.
  2018{\natexlab{a}}.
\newblock \href {https://doi.org/10.18653/v1/P18-1213} {Modeling naive
  psychology of characters in simple commonsense stories}.
\newblock In \emph{Proceedings of the 56th Annual Meeting of the Association
  for Computational Linguistics, {ACL} 2018, Melbourne, Australia, July 15-20,
  2018, Volume 1: Long Papers}, pages 2289--2299. Association for Computational
  Linguistics.

\bibitem[{Rashkin et~al.(2018{\natexlab{b}})Rashkin, Sap, Allaway, Smith, and
  Choi}]{DBLP:conf/acl/SmithCSRA18}
Hannah Rashkin, Maarten Sap, Emily Allaway, Noah~A. Smith, and Yejin Choi.
  2018{\natexlab{b}}.
\newblock \href {https://doi.org/10.18653/v1/P18-1043} {Event2mind: Commonsense
  inference on events, intents, and reactions}.
\newblock In \emph{Proceedings of the 56th Annual Meeting of the Association
  for Computational Linguistics, {ACL} 2018, Melbourne, Australia, July 15-20,
  2018, Volume 1: Long Papers}, pages 463--473. Association for Computational
  Linguistics.

\bibitem[{Rudinger et~al.(2020)Rudinger, Shwartz, Hwang, Bhagavatula, Forbes,
  Bras, Smith, and Choi}]{DBLP:conf/emnlp/RudingerSHBFBSC20}
Rachel Rudinger, Vered Shwartz, Jena~D. Hwang, Chandra Bhagavatula, Maxwell
  Forbes, Ronan~Le Bras, Noah~A. Smith, and Yejin Choi. 2020.
\newblock \href {https://doi.org/10.18653/v1/2020.findings-emnlp.418} {Thinking
  like a skeptic: Defeasible inference in natural language}.
\newblock In \emph{Proceedings of the 2020 Conference on Empirical Methods in
  Natural Language Processing: Findings, {EMNLP} 2020, Online Event, 16-20
  November 2020}, pages 4661--4675. Association for Computational Linguistics.

\bibitem[{Sakaguchi et~al.(2020)Sakaguchi, Bras, Bhagavatula, and
  Choi}]{DBLP:conf/aaai/SakaguchiBBC20}
Keisuke Sakaguchi, Ronan~Le Bras, Chandra Bhagavatula, and Yejin Choi. 2020.
\newblock \href {https://aaai.org/ojs/index.php/AAAI/article/view/6399}
  {Winogrande: An adversarial winograd schema challenge at scale}.
\newblock In \emph{The Thirty-Fourth {AAAI} Conference on Artificial
  Intelligence, {AAAI} 2020, The Thirty-Second Innovative Applications of
  Artificial Intelligence Conference, {IAAI} 2020, The Tenth {AAAI} Symposium
  on Educational Advances in Artificial Intelligence, {EAAI} 2020, New York,
  NY, USA, February 7-12, 2020}, pages 8732--8740. {AAAI} Press.

\bibitem[{Sap et~al.(2019{\natexlab{a}})Sap, Bras, Allaway, Bhagavatula,
  Lourie, Rashkin, Roof, Smith, and Choi}]{DBLP:conf/aaai/SapBABLRRSC19}
Maarten Sap, Ronan~Le Bras, Emily Allaway, Chandra Bhagavatula, Nicholas
  Lourie, Hannah Rashkin, Brendan Roof, Noah~A. Smith, and Yejin Choi.
  2019{\natexlab{a}}.
\newblock \href {https://doi.org/10.1609/aaai.v33i01.33013027} {{ATOMIC:} an
  atlas of machine commonsense for if-then reasoning}.
\newblock In \emph{The Thirty-Third {AAAI} Conference on Artificial
  Intelligence, {AAAI} 2019, The Thirty-First Innovative Applications of
  Artificial Intelligence Conference, {IAAI} 2019, The Ninth {AAAI} Symposium
  on Educational Advances in Artificial Intelligence, {EAAI} 2019, Honolulu,
  Hawaii, USA, January 27 - February 1, 2019}, pages 3027--3035. {AAAI} Press.

\bibitem[{Sap et~al.(2019{\natexlab{b}})Sap, Rashkin, Chen, Bras, and
  Choi}]{DBLP:journals/corr/abs-1904-09728}
Maarten Sap, Hannah Rashkin, Derek Chen, Ronan~Le Bras, and Yejin Choi.
  2019{\natexlab{b}}.
\newblock \href {https://doi.org/10.18653/v1/D19-1454} {Social iqa: Commonsense
  reasoning about social interactions}.
\newblock In \emph{Proceedings of the 2019 Conference on Empirical Methods in
  Natural Language Processing and the 9th International Joint Conference on
  Natural Language Processing, {EMNLP-IJCNLP} 2019, Hong Kong, China, November
  3-7, 2019}, pages 4462--4472. Association for Computational Linguistics.

\bibitem[{Sharma et~al.(2018)Sharma, Allen, Bakhshandeh, and
  Mostafazadeh}]{DBLP:conf/acl/SharmaABM18}
Rishi Sharma, James Allen, Omid Bakhshandeh, and Nasrin Mostafazadeh. 2018.
\newblock \href {https://doi.org/10.18653/v1/P18-2119} {Tackling the story
  ending biases in the story cloze test}.
\newblock In \emph{Proceedings of the 56th Annual Meeting of the Association
  for Computational Linguistics, {ACL} 2018, Melbourne, Australia, July 15-20,
  2018, Volume 2: Short Papers}, pages 752--757. Association for Computational
  Linguistics.

\bibitem[{Smith(2016)}]{Smith2016TheBO}
Tiffany~Watt Smith. 2016.
\newblock The book of human emotions.
\newblock In \emph{Little, Brown, and Company}.

\bibitem[{Socher et~al.(2013)Socher, Perelygin, Wu, Chuang, Manning, Ng, and
  Potts}]{DBLP:conf/emnlp/SocherPWCMNP13}
Richard Socher, Alex Perelygin, Jean Wu, Jason Chuang, Christopher~D. Manning,
  Andrew~Y. Ng, and Christopher Potts. 2013.
\newblock \href {https://www.aclweb.org/anthology/D13-1170/} {Recursive deep
  models for semantic compositionality over a sentiment treebank}.
\newblock In \emph{Proceedings of the 2013 Conference on Empirical Methods in
  Natural Language Processing, {EMNLP} 2013, 18-21 October 2013, Grand Hyatt
  Seattle, Seattle, Washington, USA, {A} meeting of SIGDAT, a Special Interest
  Group of the {ACL}}, pages 1631--1642. {ACL}.

\bibitem[{Vaswani et~al.(2017)Vaswani, Shazeer, Parmar, Uszkoreit, Jones,
  Gomez, Kaiser, and Polosukhin}]{DBLP:conf/nips/VaswaniSPUJGKP17}
Ashish Vaswani, Noam Shazeer, Niki Parmar, Jakob Uszkoreit, Llion Jones,
  Aidan~N. Gomez, Lukasz Kaiser, and Illia Polosukhin. 2017.
\newblock \href {http://papers.nips.cc/paper/7181-attention-is-all-you-need}
  {Attention is all you need}.
\newblock In \emph{Advances in Neural Information Processing Systems 30: Annual
  Conference on Neural Information Processing Systems 2017, December 4-9, 2017,
  Long Beach, CA, {USA}}, pages 5998--6008.

\bibitem[{Wolf et~al.(2020)Wolf, Debut, Sanh, Chaumond, Delangue, Moi, Cistac,
  Rault, Louf, Funtowicz, Davison, Shleifer, von Platen, Ma, Jernite, Plu, Xu,
  Scao, Gugger, Drame, Lhoest, and Rush}]{DBLP:conf/emnlp/WolfDSCDMCRLFDS20}
Thomas Wolf, Lysandre Debut, Victor Sanh, Julien Chaumond, Clement Delangue,
  Anthony Moi, Pierric Cistac, Tim Rault, R{\'{e}}mi Louf, Morgan Funtowicz,
  Joe Davison, Sam Shleifer, Patrick von Platen, Clara Ma, Yacine Jernite,
  Julien Plu, Canwen Xu, Teven~Le Scao, Sylvain Gugger, Mariama Drame, Quentin
  Lhoest, and Alexander~M. Rush. 2020.
\newblock \href {https://doi.org/10.18653/v1/2020.emnlp-demos.6} {Transformers:
  State-of-the-art natural language processing}.
\newblock In \emph{Proceedings of the 2020 Conference on Empirical Methods in
  Natural Language Processing: System Demonstrations, {EMNLP} 2020 - Demos,
  Online, November 16-20, 2020}, pages 38--45. Association for Computational
  Linguistics.

\bibitem[{Xu et~al.(2020)Xu, Patwary, Shoeybi, Puri, Fung, Anandkumar, and
  Catanzaro}]{DBLP:conf/emnlp/XuPSPFAC20}
Peng Xu, Mostofa Patwary, Mohammad Shoeybi, Raul Puri, Pascale Fung, Anima
  Anandkumar, and Bryan Catanzaro. 2020.
\newblock \href {https://doi.org/10.18653/v1/2020.emnlp-main.226}
  {{MEGATRON-CNTRL:} controllable story generation with external knowledge
  using large-scale language models}.
\newblock In \emph{Proceedings of the 2020 Conference on Empirical Methods in
  Natural Language Processing, {EMNLP} 2020, Online, November 16-20, 2020},
  pages 2831--2845. Association for Computational Linguistics.

\bibitem[{Zhang et~al.(2018)Zhang, Wang, and
  Liu}]{DBLP:journals/widm/ZhangWL18}
Lei Zhang, Shuai Wang, and Bing Liu. 2018.
\newblock \href {https://doi.org/10.1002/widm.1253} {Deep learning for
  sentiment analysis: {A} survey}.
\newblock \emph{Wiley Interdiscip. Rev. Data Min. Knowl. Discov.}, 8(4).

\bibitem[{Zhou et~al.(2016)Zhou, Zhang, Zhou, Zhao, and
  Geng}]{DBLP:conf/emnlp/ZhouZZZG16}
Deyu Zhou, Xuan Zhang, Yin Zhou, Quan Zhao, and Xin Geng. 2016.
\newblock \href {https://doi.org/10.18653/v1/d16-1061} {Emotion distribution
  learning from texts}.
\newblock In \emph{Proceedings of the 2016 Conference on Empirical Methods in
  Natural Language Processing, {EMNLP} 2016, Austin, Texas, USA, November 1-4,
  2016}, pages 638--647. The Association for Computational Linguistics.

\end{thebibliography}
\bibliographystyle{acl_natbib}

\clearpage

\appendix

\section{Appendix}
\label{sec:appendix}

\subsection{Baselines}
\label{appdix-baseline}

\noindent\textbf{1. GRU} \cite{DBLP:journals/corr/ChungGCB14} is a one-layer bi-GRU encodes the input text and concatenates the final time step hidden states from both directions to yield the sentence representation $h_s$.

\noindent\textbf{2. BERT}\cite{DBLP:conf/naacl/DevlinCLT19} is a standard pre-trained language model.
% To encode the input text using \textsc{BERT}, we follow the classification setup implementation described by authors.
We concatenate sentences using specific separator tokens (\texttt{[CLS]} and \texttt{[SEP]}). Finally, we take the hidden state representation of \texttt{[CLS]} in the last layer of BERT as the overall representation $h_s$ of sentence pairs;

\noindent\textbf{3. RoBERTa} \cite{DBLP:journals/corr/abs-1907-11692} is a improved robust BERT which shows state-of-the-art results in many NLP tasks. We use the hidden state representation of \texttt{<s>} as the sentence representation $h_s$.

\subsection{Implement Details}
\label{appdix-imp}

We train baseline models on 9k \textbf{\textsc{Hail}} training examples, then select hyper-parameters based on the best performing model on the dev set (2k), and then report results on the test set (2k).
The hyper-parameters of \textsc{Bart} and \textsc{Gpt}-2 is shown in Table \ref{app-1}. The hyper-parameters of \textsc{Bert} and \textsc{Roberta} is shown in Table \ref{app-2}. We use V-100 GPU to run the experiments.

\begin{table}[ht]
\centering
\scalebox{0.9}{
\begin{tabular}{p{5cm}lc}
\toprule[1pt]
Hyper-parameter                  & Value      \\
\toprule[0.5pt]
LR                               & \{1e-5, 2e-5\} \\
Batch size                       & \{16, 32, 64\}     \\
Gradient norm                    & 1.0        \\
Warm-up                          & 0.1        \\
Max. input length (\# subwords)  & 200        \\
Epochs                    & \{3, 5, 10\} \\
\toprule[1pt]
\end{tabular}}
\caption{Hyper-parameters of models based on \textsc{Bert} and \textsc{Roberta} for emotion and motivation understanding tasks.}
\label{app-1}
\end{table}

\begin{table}[ht]
\centering
\begin{tabular}{p{5cm}lc}
\toprule[1pt]
Hyper-parameter                  & Value      \\
\toprule[0.5pt]
LR                               & 1e-5  \\
$\lambda_1$                         & 1  \\
$\lambda_2$                         & 1.5  \\
Batch size                       & 32     \\
Gradient norm                    & 1.0        \\
Warm-up                          & 0.1        \\
Max. input length (\# subwords)  & 200        \\
Max. output length (\# subwords) & 60         \\
Max \# Epochs                    & 30 \\
\toprule[1pt]
\end{tabular}
\caption{Hyper-parameters of models based on \textsc{Bart} and \textsc{Gpt}-2 for action prediction task.}
\label{app-2}
\end{table}

\subsection{Data Collection}
\label{appdix-data}

To verify the effectiveness of our cognitive framework, we construct a new dataset \textbf{\textsc{Hail}} (\textbf{H}uman \textbf{A}ctivities \textbf{I}n \textbf{L}ife)  for the above four tasks by automatically extracting from the existing resource, Story Commonsense\cite{DBLP:conf/acl/KnightCSRB18}, with the help of NLP tools.

\begin{figure*}[ht]
    \centering
    \includegraphics[width=0.9\textwidth]{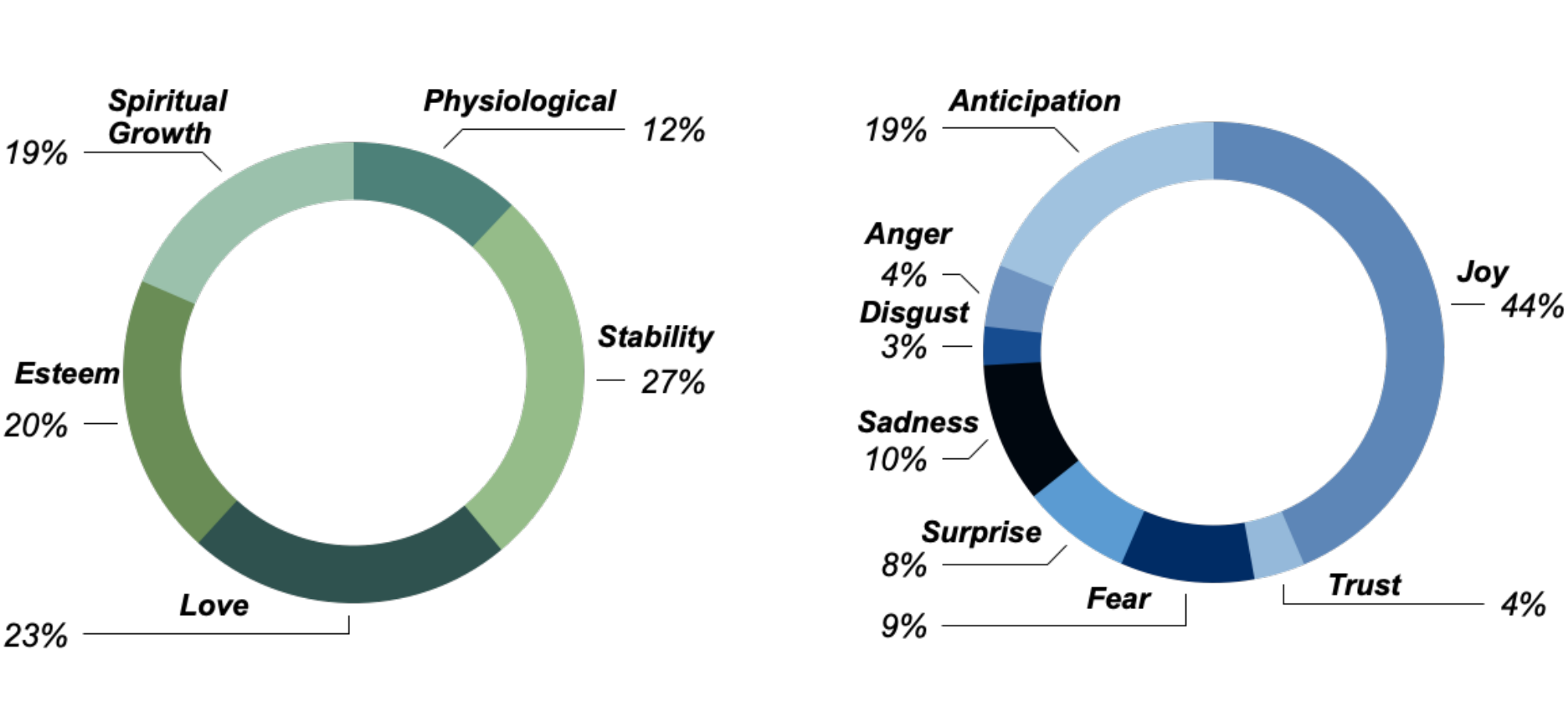}
    \caption{Data statistics of label distributions in \textbf{\textsc{Hail}}, including Human motivations (Left Pic) and Emotion Reactions (Right Pic).}
    \label{fig:data_analysis}
\end{figure*}

Our goal is to first collect training and evaluation data for the proposed four tasks. Story Commonsense dataset manually annotates human motivations and emotions of the event in daily commonsense stories. It is an important resource for studying the causality of motivations, actions, and emotions in language-based individual activities. Note that, the actors of the actions are required to have both motivations and emotion labeling in our collected data \textbf{\textsc{Hail}}. In order to obtain such \textit{(motivation, action, emotion)} samples, we utilize NLTK\footnote{http://www.nltk.org/} (a natural language processing toolkit) and design some rules. In all, we extract 13,568 examples in Story Commonsense that meet our requirements. Fig. \ref{fig:data_analysis} denotes the data statistics of label distributions in \textbf{\textsc{Hail}}, including motivations and emotions. The label distribution is relatively uniform, which is conducive to the learning of the model.

\begin{figure}[ht]
    \centering
    \includegraphics[width=0.35\textwidth]{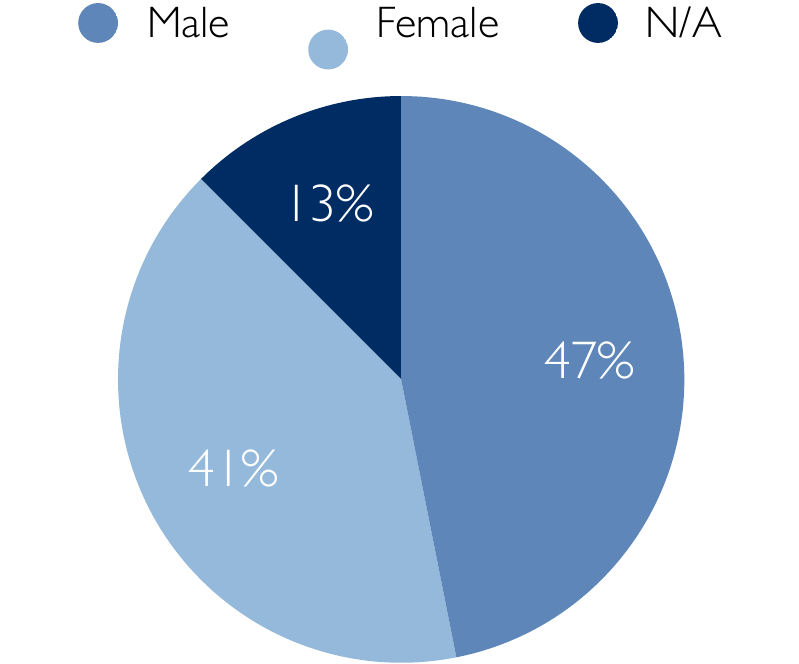}
    \caption{Analysis of gender bias of HAIL.}
    \label{fig:se}
\end{figure}

\noindent\textbf{Data Analysis} We perform analysis about the gender bias of open-text actions in \textsc{Hail}. As is shown Fig. \ref{fig:se}, our dataset have a good distribution considering the gender of individual in all actions.

This mechanism ensures that there is a clear and agreed-upon relationship between \textit{needs-action-emotion} in the story, and avoids subjectivity and ambiguity in SCT \cite{DBLP:conf/acl/SharmaABM18} and certain NLU tasks \cite{DBLP:conf/acl/NieWDBWK20}.

\begin{figure*}[ht]
    \centering
    \includegraphics[width=0.8\textwidth]{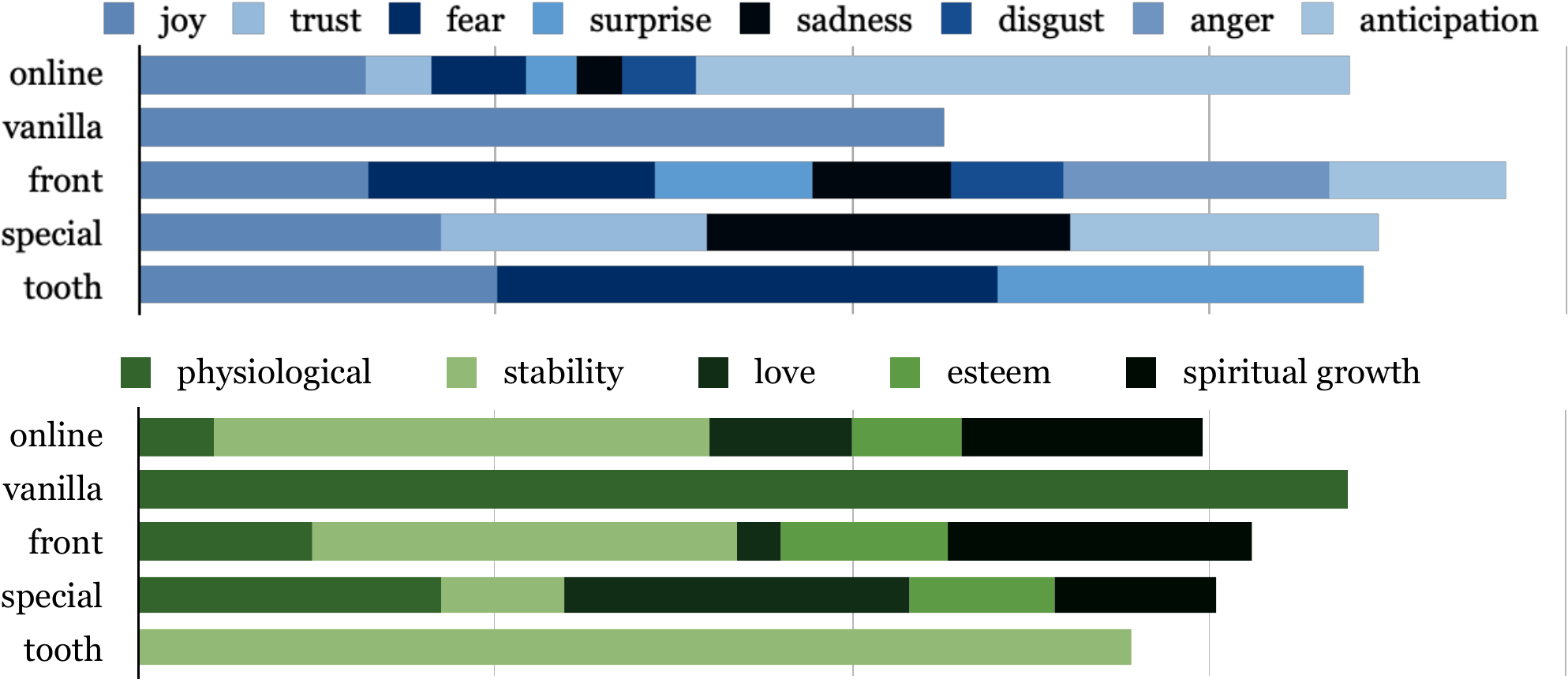}
    \caption{Examples of distribution of motivations Concept KB (below) or Emotion Concept KB (top).}
    \label{fig:vocab}
\end{figure*}

\begin{figure*}[ht]
    \centering
    \includegraphics[width=1\textwidth]{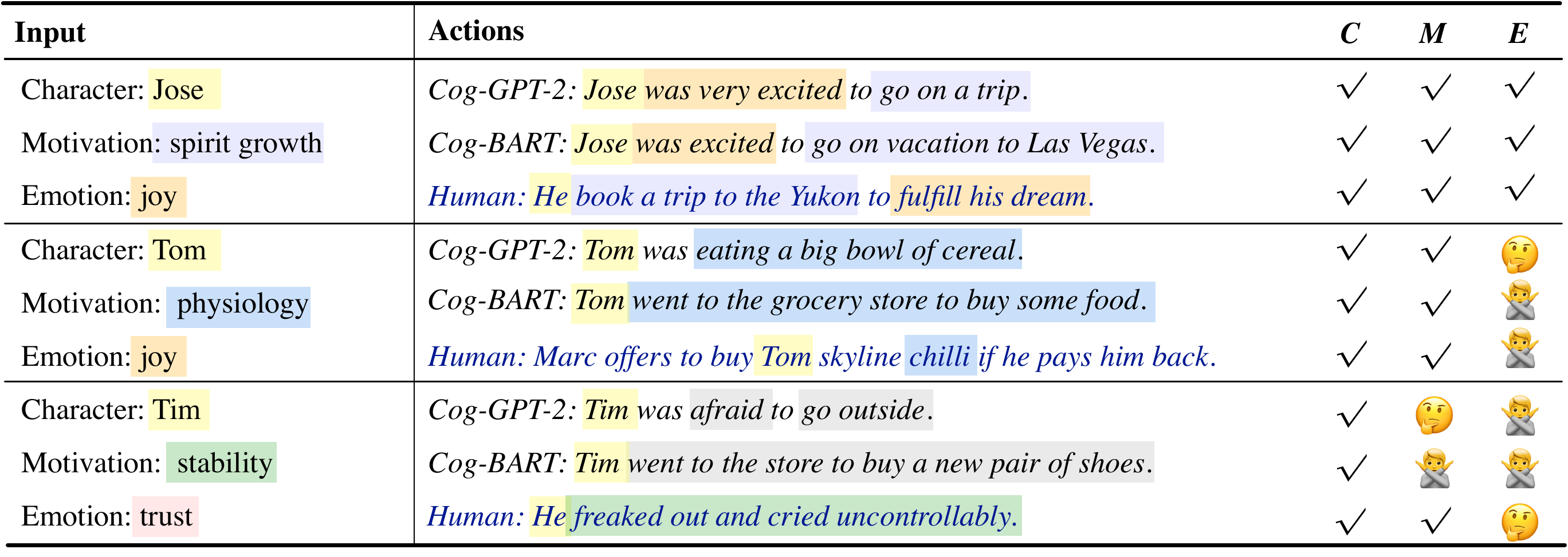}
    \caption{Case study with different inputs on action generation task. The actions are generated by \textsc{Cog-Gpt}-2, \textsc{Cog-Bart} or human writing with the specific character, motivation and emotion. $\surd$ means reasonable. \includegraphics[width=0.8em]{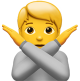} means that the generated action can not express the corresponding aspect. \includegraphics[width=0.8em]{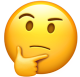} represents that the consistency is debatable. 
The colored text means generated action satisfy the aspects of mental states.}
    \label{c}
\end{figure*}

\begin{table*}[ht]
\centering
\scalebox{1}{
\begin{tabular}{cc}
\toprule[1pt]
\bf Input Prompt Template  & \bf Output Prompt Template      \\
\toprule[0.5pt]
\multicolumn{1}{p{11cm}}{[ht] C's history actions are $\_\_$ [/ht] and [mot] C has $\_\_$ motivation [/mot]} & [act] $\_\_$ [/act] \\
\toprule[1pt]
\end{tabular}}
\caption{In conditioned action generation task, input formats of \textsc{Cog-Bart} and \textsc{Cog-Gpt2}.}
\label{app-5}
\end{table*}

\subsection{Prompt Template for Input}
\label{appdix-prompt}

In emotion understanding task, a model is given history actions $\mathcal{H}$, character $\mathcal{C}$, a label of motivations $\mathcal{M}$ and a textual action $A$. 
In motivation understanding task, inputs of model are  history actions $\mathcal{H}$, character $\mathcal{C}$, a emotional label $\mathcal{E}$ and a textual action $A$. 
In conditoned action generation task, history actions $\mathcal{H}$, character $\mathcal{C}$, a label of motivations $\mathcal{M}$ and a emotional label $\mathcal{E}$ are given to the generator. 
We design simple prompt templates to expand the semantic information of the motivation and emotion labels, and also indicate the character owning the motivation and emotion. All templates for EU and MU tasks are as the following:

\noindent \texttt{C's history actions are \_\_.\\
C's motivation is \_\_ . \\ 
C's action is \_\_. \\
C's emotion is \_\_. 
}

Table \ref{app-5} shows the template for conditioned action generation.
In summary, this technique can enrich the semantic information of the labels and bring the labels with the given character's information.
Ablation studies show the effectiveness of prompt template.
%We define prompt template as $T(\cdot)$.

\subsection{Comparison of Different Inputs}
\label{appdix-case}

In order to analyze the effectiveness of our baseline models for action generation, we also perform some case studies with different inputs. The actions, with the inputs of specific history actions $\mathcal{H}$, character $\mathcal{C}$, motivation $\mathcal{M}$ or emotion $\mathcal{E}$, are generated by \textsc{Cog-Gpt}-2, \textsc{Cog-Bart} or human writing. $\surd$ means consistent to the input, while $\times$ means that the generated action can not express the corresponding aspect. The thinking face shows that the consistency of input and generated action is debatable. From the first line of Fig. \ref{app-newcase}, \textit{Jose has a spirit growth need} and \textit{Jose's emotional expectation is joy}. As demonstrated by the three actions, we can conclude that the \textsc{Cog-Gpt}-2, \textsc{Cog-Bart} based models can generate reasonable actions. Interestingly, \textsc{Cog-Bart} can guess that the destination of the trip is Las Vegas, which is competitive to human writing. In the second example, all actions can not clearly express that the emotional expectation is \textit{joy}, where \textit{a big bowl} generated by \textsc{Cog-Gpt}-2 could somewhat show the happiness of Tom. Last but not least, \textit{Tim was afraid to go outside} and \textit{Tim went to the store to buy a new pair of shoes} are plausible corresponding to the stability need. It is possible that stability need is more abstract for pre-trained language model (PLM) to understand. Besides, we can find that all actions lack the expression of \textit{trust} emotion. One reason is that the PLM based models are insensitive to emotional inputs, which is challengeable in future work.

\subsection{Knowledge Distribution of KBs}
\label{appdix-kb}

The knowledge bases can give the knowledge distribution of the motivation/emotion category according to the commonsense concepts appeared in the action, which corresponds to \textit{Knowledge Distribution}. Specifically, we use tools such as NLTK\footnote{http://www.nltk.org/} and Spacy\footnote{https://spacy.io/}, and then remove stop words and high-frequency words to extract representative commonsense concepts corresponding to the current action. 
Finally, we use each commonsense concept to retrieve the corresponding distribution in MCKB or ECKB.
In this way, the distribution of all commonsense knowledge $\{P_{c_1},P_{c_2},\dots,P_{c_n}\}$ in the current motivation/emotion category is obtained.

Base on the training set of our proposed \textbf{\textsc{Hail}}, we automatically construct knowledge bases of motivations and emotions. Examples of them are shown individually in Fig. \ref{fig:vocab}. These two KBs can be used to make prediction or assist the decision-making of the deep model, Moreover, it can also be used to evaluate or explain the forecast results.

\subsection{Future work}
Modeling the relationship among motivations, emotions and actions in human activities can allow researchers to reason the essential causes of human activities from the cognitive perspective and supply reasonable explanations. In future work, we will explore \method on various NLP downstream applications, such as intelligent dialogue, controllable text generation and public opinion analysis.

\end{document}